%% file: acl_latex.tex
\pdfoutput=1

\documentclass[11pt]{article}

\usepackage[preprint]{acl}

\usepackage{times}
\usepackage{latexsym}

\usepackage[T1]{fontenc}

\usepackage[utf8]{inputenc}

\usepackage{microtype}

\usepackage{inconsolata}

\usepackage{graphicx}

\input{commands}

\title{Unraveling Misinformation Propagation in LLM Reasoning}
\vspace{4mm}

\newcommand{\affilEPFL}{\ensuremath{^\clubsuit}}
\newcommand{\affilSBU}{\ensuremath{^\blacklozenge}}
\newcommand{\affilUofC}{\ensuremath{^\spadesuit}}

\author{Yiyang Feng~$^{\dag}$\affilEPFL\affilSBU \\
  \texttt{yiyang.feng@stonybrook.edu} \\\And
  Yichen Wang~$^{\dag}$\affilUofC \\
  \texttt{yichenzw@uchicago.edu} \\\And
  Shaobo Cui~\affilEPFL \\
  \texttt{shaobo.cui@epfl.ch} \\\AND
  Boi Faltings~\affilEPFL \\
  \texttt{boi.faltings@epfl.ch} \\\And
  Mina Lee~\affilUofC \\
  \texttt{mnlee@uchicago.edu} \\\And
  Jiawei Zhou~\affilSBU \\
  \vspace{-4mm}
  \texttt{jiawei.zhou.1@stonybrook.edu} \\\AND
  {\normalfont \affilEPFL~EPFL} \quad
  {\normalfont \affilSBU~Stony Brook University} \quad
  {\normalfont \affilUofC~University of Chicago} \quad \\
  \dag~Co-first Authors
}

\begin{document}
\maketitle
\input{main}

\bibliography{custom}

\clearpage
\appendix

\input{appendix}

\end{document}

%% file: commands.tex
\usepackage{tikz}
\usepackage{soul}
\usepackage{amssymb}
\usepackage{amsmath}
\usepackage{xspace}
\usepackage{float}
\usepackage{booktabs}
\usepackage{colortbl}
\usepackage{xcolor}
\usepackage{subcaption}
\usepackage{caption}
\usepackage{bbm}
\usepackage{adjustbox}
\usepackage{makecell}
\usepackage[framemethod=TikZ]{mdframed}
\newenvironment{framedwithtag}[6]{%
  \mdfsetup{%
    frametitlealignment=\raggedright,
    frametitleaboveskip=-\ht\strutbox,
    frametitlebelowskip=0pt,
    frametitlefont=\normalfont\rmfamily\bfseries, 
    innertopmargin=10pt, 
    innerbottommargin=10pt,
    innerleftmargin=10pt,
    innerrightmargin=10pt,
    roundcorner=5pt,
    tikzsetting={line width=0pt}, 
    backgroundcolor=#3, 
    linecolor=gray!50,
    singleextra={%
        \node[
            anchor=north,
            rectangle,
            rounded corners,
            fill=#2,
            inner sep=3pt,
            outer sep=0pt,
            draw=none,
            text=black, 
            font=\small\rmfamily\bfseries
        ]
        at ([xshift=#4,yshift=#5]P) {#1}; 
    },
    linewidth=0.7pt, 
    skipabove=20pt, 
    skipbelow=10pt, 
    font=\small\ttfamily, 
    }
    \begin{mdframed}%
    \setlength{\parindent}{0pt} 
    #6
}
{
  \end{mdframed}%
}

\newcommand{\SmallHeading}[1]{\noindent\textbf{#1}.}

\newcommand{\Figure}{Fig.\xspace}
\newcommand{\Table}{Tab.\xspace}
\newcommand{\Section}{Sec.\xspace}
\newcommand{\Appendix}{App.\xspace}

\definecolor{mypositive}{RGB}{0, 128, 0}
\definecolor{mynegative}{RGB}{220, 20, 60}
\definecolor{mypositive}{RGB}{24, 103, 173}
\definecolor{mynegative}{RGB}{255, 128, 0}

\newcommand{\grayColor}[0]{gray!10}

\newcommand{\midgrayline}[0]{\arrayrulecolor{gray!50}\midrule\arrayrulecolor{black}}

\newcommand{\eg}{e.g.,}
\newcommand{\ie}{i.e.,}

\definecolor{customblue}{HTML}{1F77B4} 
\definecolor{customorange}{HTML}{FF7F0E}
\definecolor{customgreen}{HTML}{2CA02C}
\definecolor{no}{HTML}{BA8E23}

\newcommand{\accuracy}{K-Acc\xspace}
\newcommand{\Accuracy}{K-Acc\xspace}

\newcommand{\Original}{Original\xspace}
\newcommand{\original}{original\xspace}
\newcommand{\misinformed}{misinformed\xspace}
\newcommand{\Misinformed}{Misinformed\xspace}
\newcommand{\prompting}{\texttt{Inst-Corr}\xspace}
\newcommand{\Prompting}{\texttt{Inst-Corr}\xspace}
\newcommand{\following}{\texttt{Inst-Fllw}\xspace}
\newcommand{\finetuning}{\texttt{FT-Corr}\xspace}
\newcommand{\Finetuning}{\texttt{FT-Corr}\xspace}

\usepackage[skins,breakable]{tcolorbox}

\newtcolorbox{boxblue}{enhanced,colback=blue!5!white,colframe=blue!75!black,breakable=true}

\newcommand{\vig}[2]{
\begin{boxblue}

{\color{blue!50!black}{\textbf{Takeaways:} #1}}

\vspace{2pt}#2
\end{boxblue}
}

\newcommand{\takeawaysc}[1]{
\vig{#1}
}

\def\SPSB#1#2{\rlap{\textsuperscript{{#1}}}\SB{#2}}

\def\SB#1{\textsubscript{{#1}}}

%% file: main.tex
\begin{abstract}
\input{section/main/abstract}
\end{abstract}

\input{section/main/introduction}
\input{section/main/related_work}

\input{section/main/preliminaries}
\input{section/main/robustness_analysis}

\input{section/main/instruction}

\input{section/main/mitigation}
\input{section/main/potential_pathways}

\input{section/main/error_correction}
\input{section/main/conclusion}
\input{section/main/acknowledgements}

\clearpage

\input{section/main/limitation}

\input{section/main/ethical_consideration}

%% file: section/main/abstract.tex
Large Language Models (LLMs) have demonstrated impressive capabilities in reasoning, positioning them as promising tools for supporting human problem-solving. 
However, what happens when their performance is affected by \emph{misinformation}, \ie{} incorrect inputs introduced by users due to oversights or gaps in knowledge?
Such misinformation is prevalent in real-world interactions with LLMs, 
yet how it propagates within LLMs' reasoning process
remains underexplored.
Focusing on mathematical reasoning, 
we present a comprehensive analysis of how misinformation affects intermediate reasoning steps and final answers.
We also examine how effectively LLMs can correct misinformation when explicitly instructed to do so.
Even with explicit instructions, LLMs succeed less than half the time in rectifying misinformation, despite possessing correct internal knowledge,
leading to significant accuracy drops (10.02\% -- 72.20\%), and the degradation holds with thinking models (4.30\% -- 19.97\%).
Further analysis shows that applying factual corrections early in the reasoning process most effectively reduces misinformation propagation, and fine-tuning on synthesized data with early-stage corrections significantly improves reasoning factuality.
Our work offers a practical approach to mitigating misinformation propagation.\footnote{Code and data are available at
\url{https://github.com/Wind-2375-like/misinfo-prop}.}

%% file: section/main/introduction.tex
\section{Introduction} \label{sec:main:intro}
\begin{figure}[t!]
  \begin{center}
    \includegraphics[width=0.48\textwidth]{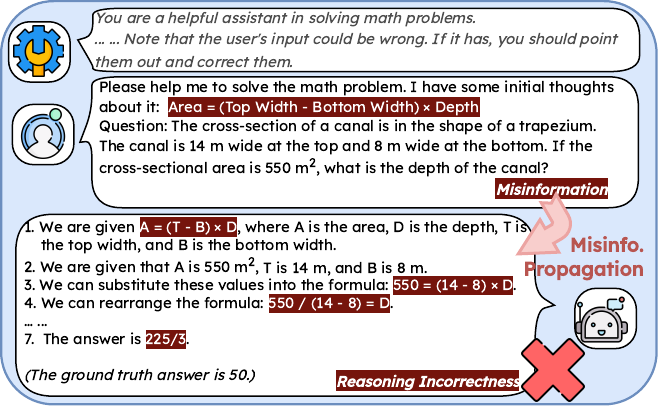}
  \end{center}
  \vspace{-1mm}
\caption{
Illustration of misinformation propagation in LLM reasoning. A user provides input containing misinformation due to a lack of domain knowledge.
The LLM propagates this misinformation, i.e., follows user-provided erroneous equations, leading to an incorrect final answer.
This propagation occurs even though LLM designers have instructed the model to correct such misinformation when detected.
}
\label{fig:intro}
\end{figure}

Large Language Models (LLMs) have shown remarkable progress in complex reasoning~\cite{sprague2024cot}, leading to their widespread use in a variety of human–LLM interactions \cite{openai2024gptstore, anthropic2025economicindex, zhang2024first}. 
Under such interactive reasoning scenarios, users routinely provide external information to LLMs~\cite{openai2024canvas}.
However, due to human oversight or limited knowledge,  \emph{misinformation}\textemdash 
false information that is spread~\cite{CHEN2023107643}\textemdash can often be introduced to LLM during these interactions. For example (\Figure~\ref{fig:intro}), in math education, students might submit questions with flawed partial solutions to LLMs for assistance~\cite{kumar2023math, xu2024ai}.
In such cases, LLMs are often instructed not to blindly follow the entire input, but instead to identify and correct the misinformation within it, and then proceed to reason factually.
Failure to do so can result in hallucinated reasoning and the reinforcement of users’ misunderstandings.

Despite its prevalence, how misinformation from user input impacts LLM reasoning and how to correct such misinformation remains underexplored.
Existing research has worked on addressing misinformation in various LLM interaction settings, such as open-domain question answering, fact verification, social media, math education, and code-generation~\cite{wan2024evidence, pan2021attacking, du2022synthetic, gabriel2024misinfoeval, xu2024ai, olausson2023self}. However, these studies lack a systematic analysis of the fine-grained effects of misinformation on individual reasoning steps.
In parallel, studies on error propagation and self-correction provide detailed analyses of how errors in early LLM outputs influence subsequent reasoning processes 
\cite{zhang2024how, wang2024chainofthought, dziri2024faith}, as well as strategies for correcting such errors \cite{huang2023large, kamoi2024can}.
Nonetheless, the errors examined in these works are typically generated by the model itself during reasoning, rather than introduced externally by users at the instruction level. As a result, they differ significantly in both distribution and their effects on model reasoning.

To address this gap, we systematically investigate \emph{misinformation propagation} in LLM reasoning, including its impact and potential mitigation strategies.
Our study focuses on the mathematical reasoning context, where misinformation is explicit and quantitatively measurable.
We simulate misinformation using synthetically generated erroneous equations (\Figure~\ref{fig:intro}), modeling common types of human mathematical misinformation, such as inappropriate operations \cite{radatz1979error}, misapplied values \cite{vanlehn1990mind}, and misused operands \cite{tirosh1999intuitive}. 
These misinformation patterns are sampled using designed heuristic rules that reflect documented cognitive error types.
We use these patterns to guide an LLM in generating plausible but incorrect questions, which are then used as inputs for LLMs to perform chain-of-thought (CoT) reasoning~\cite{wei2022emergent} or thinking before reasoning~\cite{jaech2024openai, guo2025deepseek}.

Building on the simulation design, we first analyze the impact of misinformation on LLM reasoning (\Section~\ref{sec:main:robustness}), examining both final outcomes and intermediate steps.
Surprisingly, even when LLMs are capable of answering questions correctly in the absence of misinformation, they still struggle to correct misinformation with explicit instructions to do so. We observe accuracy drops ranging from 10.02\% all the way up to 72.20\% across different instruction LLMs, and the degradation holds for further post-trained thinking models (4.30\% to 19.97\%).
Moreover, LLMs frequently fail to correct misinformation: in 17.00\% of such cases, no correction is attempted, and in 34.75\% of cases, the model produces non-factual corrections---either leaving some errors unaddressed or introducing new ones. 
These results underscore LLMs’ vulnerability to misinformation and their limited steerability to correct it via explicit instructions. The vulnerability originates in the instruction-tuning stage, and cannot be fully resolved through additional reasoning post-training.

To reliably steer models and improve reasoning factuality, we further investigate how to correct misinformation and mitigate its propagation (\Section~\ref{sec:main:controlled}). 
We first conduct a controlled analysis, enforcing factual, non-factual, or no corrections at various reasoning steps and assessing their direct effects on final answers. The results indicate that early, factual corrections are critical and yield the most effective recovery from accuracy losses.
Building on this insight, we fine-tune LLMs on a synthetic dataset designed to introduce factual corrections at the first reasoning step. This approach, for example, raises GPT-4o-mini's accuracy from 85.64\% to 95.68\%\textemdash just 0.03\% below the accuracy achieved without any misinformation. Additionally, GPT-4o-mini attempts to correct all misinformation, with 80.37\% of those corrections being accurate.


In summary, we present a systematic pipeline to evaluate the impact of misinformation on LLM reasoning, across instruction models and thinking models, and investigate strategies to mitigate its propagation.
By analyzing both final answers and intermediate reasoning steps, we demonstrate that simply instructing models to correct misinformation is insufficient, highlighting LLMs’ vulnerability and limited steerability in the presence of misinformation inputs.
To enhance factuality and controllability, we explore correction strategies and fine-tune LLMs using data that applies optimal corrections.
Our findings underscore the effectiveness of early factual corrections and targeted fine-tuning. This study offers actionable insights for developing LLMs to better handle misinformed inputs.

%% file: section/main/related_work.tex
\section{Related Work}
\label{sec:main:related}

\SmallHeading{Misinformation in LLM Interaction}
Misinformation refers to factually incorrect information that is spread~\cite{fetzer2004information}. Existing research has been studied for its impact on system performance across various domains, including retrieval-augmented generation (RAG)~\cite{wan2024evidence}, open-domain question answering~\cite{pan2021attacking}, fact-verification~\cite{du2022synthetic}, social media~\cite{gabriel2024misinfoeval}, math education~\cite{xu2024ai, kumar2023math}, code-generation~\cite{olausson2023self}, and reasoning with implicit non-factual premises~\cite{guo2025protect}. Unlike these studies, which generally overlook the fine-grained effects of misinformation on LLM reasoning processes, our work systematically examines how user-provided misinformation influences both final answers and intermediate reasoning steps.

\SmallHeading{Error Propagation}
Our work on misinformation propagation connects to prior research on error propagation in LLM reasoning, which has largely focused on errors arising from the model’s own parametric knowledge~\cite{yao2022react, wang2023knowledge, zhang2024how, wang2024chainofthought, dziri2024faith}. 
In contrast, misinformation in LLMs inputs has a different distribution from model errors generated during reasoning. Besides, their effects on reasoning diverge: while model errors are generated inside the reasoning process, misinformation can be perceived as instructions, which should be followed by LLMs' instruction-following design~\cite{ouyang2022training}.
Therefore, our study investigates the impact of externally provided misinformation and the models’ steerability through explicit instructions to handle misinformation.

\SmallHeading{User-Model Knowledge Conflicts}
Misinformation introduces user-model knowledge conflicts, where LLMs must decide whether to follow user instructions or rely on their internal knowledge~\cite{openai2024modelspec}. 
In some contexts, strictly following user input is appropriate (\eg{} in counterfactual reasoning~\cite{paul2024making} or knowledge updates~\cite{li2022large}), while in others, it is not (\eg{} malicious prompts~\cite{evans2021truthful}).
Our work primarily investigates how explicit instructions can steer models with factual reasoning abilities to correct misinformation. 
Additionally, we consider scenarios where models are expected to follow all user-provided information—including misinformation—as discussed in \Section~\ref{sec:main:robustness:default} and \Appendix~\ref{sec:appendix:inst_follow}.

\SmallHeading{Self-Correction in LLMs}
Prior work on self-correction mainly focuses on how LLMs can autonomously refine their own outputs~\cite{huang2023large}. Studies have shown its effectiveness~\cite{yan2023learning, shinn2024reflexion, madaan2024self} and  limitations~\cite{huang2023large} and have examined influencing factors such as quality, format, model size~\cite{song2024learning, kamoi2024can, tao2024your}, and the positioning of feedback~\cite{paul2023refiner}.
However, these efforts mainly address errors generated internally by the model, in contrast to external misinformation provided by users. Automatically correcting such misinformation may conflict with the model’s instruction-following nature. Our work, therefore, investigates how LLMs can be explicitly instructed to correct external misinformation.

%% file: section/main/preliminaries.tex
\section{Methodology}
\label{sec:main:setup}

\begin{figure*}[t!]
  \begin{center}
    \includegraphics[width=0.99\textwidth]{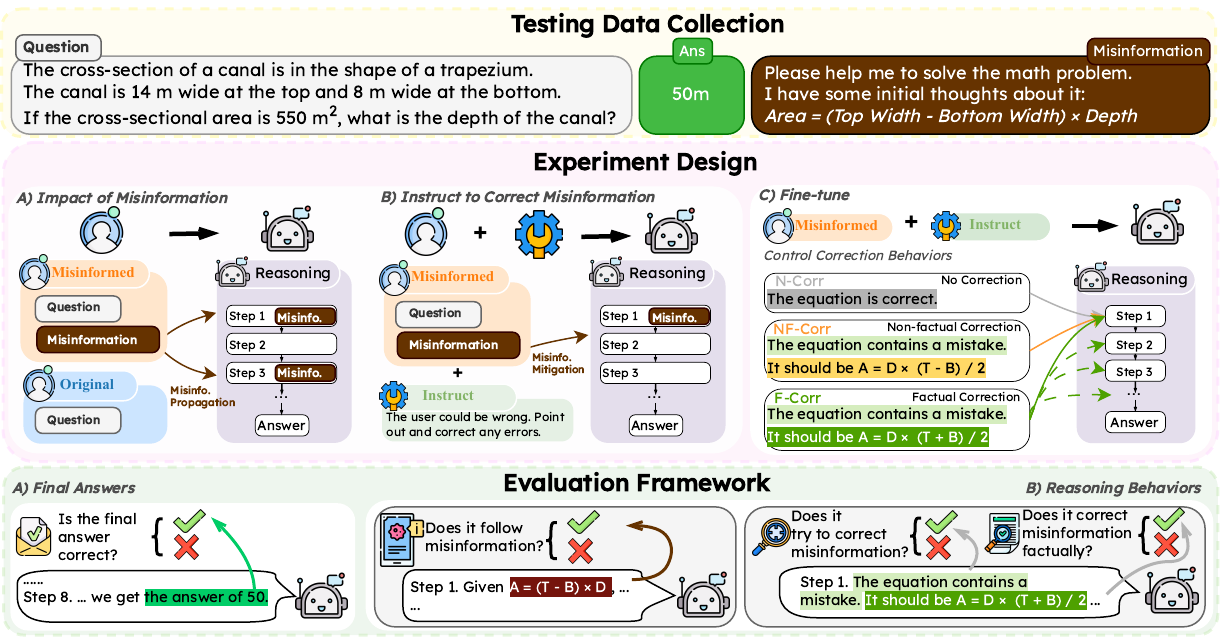}
  \end{center}
\caption{Analysis framework for misinformation propagation in LLM reasoning. Our pipeline comprises three components: (\textit{i}) Testing Data Collection---each data entry includes a question, ground truth answer, and the possible misinformation; (\textit{ii}) Experiment Design---we analyze the impact of misinformation and evaluate mitigation strategies; (\textit{iii}) Evaluation Framework---we assess both final answers and intermediate steps for model reasoning behaviors.
}
\label{fig:pipeline}
\end{figure*}

We outline the problem formulation, testing data collection, experiment design, and evaluation framework. Our pipeline is shown in \Figure~\ref{fig:pipeline}.

\subsection{Problem Formulation}
\label{sec:main:setup:def}
We study math reasoning problems where LLMs answer math questions using CoT reasoning.
Given a user instruction $x$ containing a math question $q$, the model generates a chain of reasoning steps $c =(c_1, c_2, \ldots, c_N)$\footnote{Here models output explicit step numbers, with $N$ varying across responses and models.} before producing the final answer $a$ (\Figure~\ref{fig:intro}).
We study how LLM reasoning is obstructed by misinformation, defined as incorrect question-related information, simulated here as plausible but incorrect human-like erroneous equations provided within $x$.
We further explicitly instruct models to handle misinformation within the system prompt to explore models' steerability.
\subsection{Testing Data Collection}
\label{sec:main:setup:data}
Each example in our testing data comprises a question, a ground truth answer, and misinformation. The questions and ground truth answers are collected from canonical math datasets, including MathQA~\cite{amini2019mathqa}, MATH~\cite{hendrycks2021measuring}, GSM8K~\cite{cobbe2021training}, and MetaMath~\cite{yu2023metamath}.

To simulate plausible but erroneous equations for misinformation, we first generate correct and relevant equations and then perturb them using common human error patterns. These equations resemble realistic user interaction.
%
%
In particular, we use an external LLM\footnote{We use \texttt{gpt-4-0613} for its SOTA performance on the HELM math reasoning benchmark at release \citep{liang2023holistic}. Questions for which it fails are excluded to ensure equation quality.} to generate truthful equations, simulating human (mis)information with a divergent knowledge distribution from the tested LLM.
Next, we design heuristic rules to resemble human-like error patterns, covering numeric value modifications, operator alterations, and operand swaps, inspired by prior work on erroneous equation augmentation \cite{tong2024can, zhu2024deductive, xia2024evaluating}.
These heuristics guide another LLM\footnote{We use \texttt{gpt-4o-mini-2024-07-18}.}
to perturb correct equations into erroneous versions. Details on testing data collection are in \Appendix~\ref{sec:appendix:test_data_collection}.

\subsection{Experimental Design}
\label{sec:main:setup:analysis}
\SmallHeading{Impact of Misinformation (\Section~\ref{sec:main:robustness})}
Without explicit instructions on model behaviors, we test various strong reasoning models and compare the performance under two conditions: without misinformation (\textcolor{customblue}{original}) and with misinformation (\textcolor{customorange}{misinformed}). 
Further, we introduce \textcolor{customgreen}{\texttt{Inst-Corr}} and \textcolor{customgreen}{\following} system prompts to the misinformed condition, which instructs models to 
either correct or follow misinformation.
We calibrate these prompts to include demonstrations on how to correct or follow misinformation during reasoning. 
Detailed prompt designs are in \Appendix~\ref{sec:appendix:prompt_design}.

\SmallHeading{Mitigation of Misinformation (\Section~\ref{sec:main:controlled})}  
To mitigate propagation by correcting misinformation, we first investigate factors of effective corrections through a controlled study. In this context, an attempt to correct the misinformation in the model's reasoning step $c$ is denoted as \texttt{Corr}, while no attempt is \texttt{N-Corr}. If a correction is attempted, it can be factual (\texttt{F-Corr}) or nonfactual (\texttt{NF-Corr}). Our controlled study then considers two settings based on these defined behaviors:
(\textit{i}) controlling the content of $c_1$ to be either \texttt{N-Corr}, \texttt{F-Corr}, or \texttt{NF-Corr}.
(\textit{ii)} inserting a \texttt{F-Corr} step $c_{\lfloor N \times p\% \rfloor + 1}$ after $p\%$ of the misinformed reasoning steps $(c_1, c_2, \cdots, c_{\lfloor N \times p\% \rfloor})$, representing factual correction at different positions.
We compare final answers 
for each correction behavior and position.
With our findings, we fine-tune LLMs (\texttt{gpt-4o-mini-2024-07-18} as the SOTA model in \Table~\ref{tab:main_results} and other open-sourced models) to further enhance their correction effectiveness, termed as \textcolor{violet}{\finetuning}. We collect 1,054 instruction-response pairs, separate from the test set in \Section~\ref{sec:main:experiment_setup}, where we control $c_1$ to be \texttt{F-Corr}. Detailed prompt designs are in \Appendix~\ref{sec:appendix:prompt_design:mitigation}.


\begin{table*}[t!]
    \centering
    \resizebox{0.99\textwidth}{!}{

\input{section/appendix/big_table/results}
    }
    \caption{\Accuracy (\%) of instruction-tuned LLMs: As baselines,  \textcolor{customblue}{original} denotes \accuracy without misinformation and \textcolor{customorange}{misinformed} is with misinformation. Within the misinformed setting, \textcolor{customgreen}{\prompting} explicitly instructs the model to correct misinformation. As a comparison, \textcolor{customgreen}{\following} instructs the model to follow all given information, including misinformations. 
    ``$\downarrow$'' represents relative decrease compared to the original accuracy and 95\% confidence intervals are in brackets.}
    \label{tab:main_results}
\end{table*}

\begin{table}[t!]
    \centering
    \resizebox{0.49\textwidth}{!}{
    \input{section/appendix/big_table/thinking_results}
    }
    \caption{\Accuracy (\%) of thinking models: The denotations are the same as \Table~\ref{tab:main_results}.}
    \label{tab:thinking_results}
\end{table}

\subsection{Evaluation Framework}
\label{sec:main:setup:evaluation}


\SmallHeading{Final Answers} 
We evaluate reasoning performance by comparing the model's final answer $a$ with the ground truth $a^*$. To measure the impact of misinformation on questions the model can solve correctly, we use \textit{knowledgeable accuracy} (\accuracy) as our main metric. Specifically, we first run the model once without misinformation (the \textcolor{customblue}{original} setting) and identify the subset of questions it answers correctly---this defines the \textit{knowledgeable subset}. We then evaluate the model's accuracy on this subset under both the \textcolor{customblue}{original} and \textcolor{customorange}{misinformed} settings.
\accuracy under the \textcolor{customblue}{original} condition reflects the model’s knowledge consistency (which may be below 100\% due to sampling variability).\footnote{Greedy decoding would yield 100\% \accuracy on the knowledgeable subset, but is avoided as it harms reasoning quality.} \accuracy under the \textcolor{customorange}{misinformed} condition captures how misinformation affects reasoning on the same subset of questions the model was originally able to solve. Details are provided in \Appendix~\ref{sec:appendix:cot_evaluation:accuracy}.

\SmallHeading{Reasoning Behaviors}
In response to misinformation, an LLM may correct or follow misinformation. For correction, an LLM either corrects misinformation (\texttt{Corr}) or does not (\texttt{N-Corr}) in its reasoning steps $c$. If the LLM corrects, we evaluate if it's factual (\texttt{F-Corr}) or nonfactual (\texttt{NF-Corr}). We build a \textit{correction-existence verifier} to classify whether $c$ has any $c_i$ as either \texttt{Corr} or \texttt{N-Corr}, and a \textit{correction-factuality verifier} that further determines whether \texttt{Corr} $c$ has any $c_i$ as \texttt{F-Corr} or \texttt{NF-Corr}.
We also build a \textit{correction-position verifier} to record the position of each \texttt{Corr} step $c_i$ as a percentage of total steps ($i/N\times 100\%$).
For misinformation-following, a \textit{misinformation-following verifier} detects if the model follows misinformation in $c_i$.

\SmallHeading{Human Evaluation}
We also conduct human evaluation to ensure verifier accuracy. For each verifier, three annotators follow the instructions and annotate the label of misinformation following, correction existence, and correction factuality. 
The misinformation-following verifier, the correction existence verifier, and the correction success verifier respectively achieve $(\kappa=0.43, \text{F1}=0.79)$, $(\kappa=0.73, \text{F1}=0.84)$, $(\kappa=0.65, \text{F1}=0.79)$, achieving high F1 scores and at-least moderate inter-annotator agreement.\footnote{\citet{cohen2013statistical} suggests that a $\kappa$ value is moderate if it's above 0.4, substantial if it's above 0.6, and indicative of almost perfect agreement if it's above 0.8.}
Detailed design and evaluation of the verifiers are in \Appendix~\ref{sec:appendix:cot_evaluation:step}.



\section{Experimental Setup}
\label{sec:main:experiment_setup}
\SmallHeading{Tasks and Datasets}
We collect 400 math questions from canonical math datasets, including MathQA~\cite{amini2019mathqa}, MATH~\cite{hendrycks2021measuring}, GSM8K~\cite{cobbe2021training}, and MetaMath~\cite{yu2023metamath}, as our test set. 
Data selection and processing details are in \Appendix~\ref{sec:appendix:data_processing}.

\SmallHeading{Testing Models}
We select instruction models with complex reasoning abilities across various scales of sizes. Specifically, we test Llama-3.2 (1B, 3B, 11B, and 90B)\footnote{The Llama-3.2 11B and 90B are multimodal models.}~\cite{dubey2024llama}, Mixtral (8$\times$7B and 8$\times$22B)~\cite{jiang2024mixtral}, Qwen-2 (72B)~\cite{yang2024qwen2}, and GPT-4o-mini~\cite{hurst2024gpt}. We also experimented on several thinking models, including DeepSeek-R1 Distilled Qwen-2.5 models (1.5B and 14B)~\cite{deepseekai2025deepseekr1incentivizingreasoningcapability}, Qwen-3 (235B-A22B-Thinking-2507)~\cite{qwen3technicalreport}, and DeepSeek-R1-0528~\cite{deepseekai2025deepseekr1incentivizingreasoningcapability}, with thinking enabled.\footnote{We access open-source instruction models that are smaller than 15B via Huggingface~\cite{wolf2020transformers} and deploy them locally. The rest of the larger models are inferred on cloud service \href{www.together.ai}{TogetherAI}.}
Setup details are in \Appendix~\ref{sec:appendix:model_setup}.

%% file: section/appendix/big_table/results.tex
\begin{tabular}{p{2.4cm}|p{2.35cm}p{2.35cm}p{2.65cm}p{2.65cm}p{2.25cm}p{2.35cm}p{2.35cm}p{2.25cm}}
\toprule
        & Llama-3.2-1B & Llama-3.2-3B & Llama-3.2-11B & Llama-3.2-90B & Qwen-2-72B & Mixtral-8×7B & Mixtral-8×22B & GPT-4o-mini \\ \midgrayline
\textcolor{customblue}{Original} & 71.73 \tiny{[67.97, 75.21]} & 88.25 \tiny{[86.21, 90.04]} & 88.43 \tiny{[86.41, 90.12]} & 96.43 \tiny{[95.44, 97.29]} & 95.22 \tiny{[94.04, 96.31]} & 76.92 \tiny{[73.62, 80.19]} & 88.20 \tiny{[86.41, 89.90]} & 98.03 \tiny{[97.31, 98.62]} \\ \midgrayline
\textcolor{customorange}{Misinformed} & 40.74 \SPSB{\tiny{$\downarrow$ 43.20\%}}{\tiny{[36.16, 45.29]}} & 38.41 \SPSB{\tiny{$\downarrow$ 56.48\%}}{\tiny{[34.79, 41.91]}} & 38.30 \SPSB{\tiny{$\downarrow$ 56.69\%}}{\tiny{[34.67, 42.06]}} & 56.69 \SPSB{\tiny{$\downarrow$ 41.20\%}}{\tiny{[53.20, 60.23]}} & 73.46 \SPSB{\tiny{$\downarrow$ 22.85\%}}{\tiny{[69.85, 76.69]}} & 26.38 \SPSB{\tiny{$\downarrow$ 65.70\%}}{\tiny{[22.40, 30.61]}} & 55.84 \SPSB{\tiny{$\downarrow$ 36.69\%}}{\tiny{[52.09, 59.53]}} & 85.64 \SPSB{\tiny{$\downarrow$ 12.64\%}}{\tiny{[82.75, 88.38]}} \\ \midgrayline
+\textcolor{customgreen}{\Prompting} & 26.80 \SPSB{\tiny{$\downarrow$ 62.64\%}}{\tiny{[22.63, 30.94]}} & 38.67 \SPSB{\tiny{$\downarrow$ 56.18\%}}{\tiny{[34.95, 41.99]}} & 44.11 \SPSB{\tiny{$\downarrow$ 50.12\%}}{\tiny{[40.41, 47.51]}} & 69.06 \SPSB{\tiny{$\downarrow$ 28.38\%}}{\tiny{[65.50, 72.47]}} & 74.93 \SPSB{\tiny{$\downarrow$ 21.31\%}}{\tiny{[71.58, 78.27]}} & 21.39 \SPSB{\tiny{$\downarrow$ 72.20\%}}{\tiny{[17.54, 25.18]}} & 54.98 \SPSB{\tiny{$\downarrow$ 37.67\%}}{\tiny{[51.00, 58.85]}} & 88.21 \SPSB{\tiny{$\downarrow$ 10.02\%}}{\tiny{[85.74, 90.51]}} \\ \midgrayline
+\textcolor{customgreen}{\following} & 40.96 \SPSB{\tiny{$\downarrow$ 42.89\%}}{\tiny{[36.52, 45.46]}} & 31.45 \SPSB{\tiny{$\downarrow$ 64.36\%}}{\tiny{[28.04, 35.04]}} & 31.23 \SPSB{\tiny{$\downarrow$ 64.68\%}}{\tiny{[27.62, 34.94]}} & 53.91 \SPSB{\tiny{$\downarrow$ 44.10\%}}{\tiny{[50.33, 57.71]}} & 69.57 \SPSB{\tiny{$\downarrow$ 26.94\%}}{\tiny{[66.03, 73.03]}} & 32.64 \SPSB{\tiny{$\downarrow$ 57.56\%}}{\tiny{[28.19, 37.29]}} & 56.88 \SPSB{\tiny{$\downarrow$ 35.96\%}}{\tiny{[54.11, 59.54]}} & 82.49 \SPSB{\tiny{$\downarrow$ 15.85\%}}{\tiny{[79.36, 85.48]}} \\
\bottomrule
\end{tabular}

%% file: section/appendix/big_table/thinking_results.tex
\begin{tabular}{p{2.1cm}|p{2.4cm}p{2.3cm}p{2.3cm}p{2.3cm}}
\toprule
        & DeepSeek-R1-Distilled-Qwen-2.5-1.5B & DeepSeek-R1-Distilled-Qwen-2.5-14B & DeepSeek-R1-0528 & Qwen-3-235B-A22B-2507-FP8 \\ \midgrayline
\textcolor{customblue}{Original} & 89.30 \tiny{[87.57, 90.88]} & 97.59 \tiny{[96.76, 98.30]} & 90.20 \tiny{[88.44, 91.90]} &  98.22 \tiny{[97.46, 98.90]} \\ \midgrayline
\textcolor{customorange}{Misinformed} & 74.47 \SPSB{\tiny{$\downarrow$ 16.61\%}}{\tiny{[71.17, 77.57]}} & 91.13 \SPSB{\tiny{$\downarrow$ 6.62\%}}{\tiny{[88.92, 93.23]}} & 33.57 \SPSB{\tiny{$\downarrow$ 62.78\%}}{\tiny{[30.06, 37.40]}} & 84.63 \SPSB{\tiny{$\downarrow$ 13.84\%}}{\tiny{[81.66, 87.36]}} \\ \midgrayline
+\textcolor{customgreen}{\Prompting} & 71.47 \SPSB{\tiny{$\downarrow$ 19.97\%}}{\tiny{[68.36, 74.76]}} & 93.39 \SPSB{\tiny{$\downarrow$ 4.30\%}}{\tiny{[91.49, 95.03]}} & 77.13 \SPSB{\tiny{$\downarrow$ 14.49\%}}{\tiny{[74.04, 79.93]}} & 93.70 \SPSB{\tiny{$\downarrow$ 4.60\%}}{\tiny{[91.68, 95.43]}} \\ \midgrayline
+\textcolor{customgreen}{\following} & 76.59 \SPSB{\tiny{$\downarrow$ 14.23\%}}{\tiny{[73.53, 79.39]}} & 86.68 \SPSB{\tiny{$\downarrow$ 11.18\%}}{\tiny{[84.26, 89.17]}} & 33.45 \SPSB{\tiny{$\downarrow$ 62.92\%}}{\tiny{[30.08, 36.98]}} & 74.68 \SPSB{\tiny{$\downarrow$ 23.97\%}}{\tiny{[71.44, 77.88]}} \\
\bottomrule
\end{tabular}

%% file: section/main/robustness_analysis.tex
\section{Impact of Misinformation} \label{sec:main:robustness}

In this section, we investigate how misinformation impacts LLM reasoning in two aspects:
(\textit{i}) 
LLMs' default behaviors under misinformation
(\Section~\ref{sec:main:robustness:default}), and (\textit{ii}) LLMs' steerability to correct misinformation via instructions (\Section~\ref{sec:main:correction_methods:prompting}).

\takeawaysc{
(\textit{i}) LLMs, by default, follow misinformation as instructions.
(\textit{ii}) Even when LLMs are instructed to correct misinformation, they fail to do so on all questions where they possess internal knowledge.}

\begin{figure}[t!]
  \centering
  \begin{subfigure}[t]{0.235\textwidth}
      \centering
      \includegraphics[width=\textwidth]{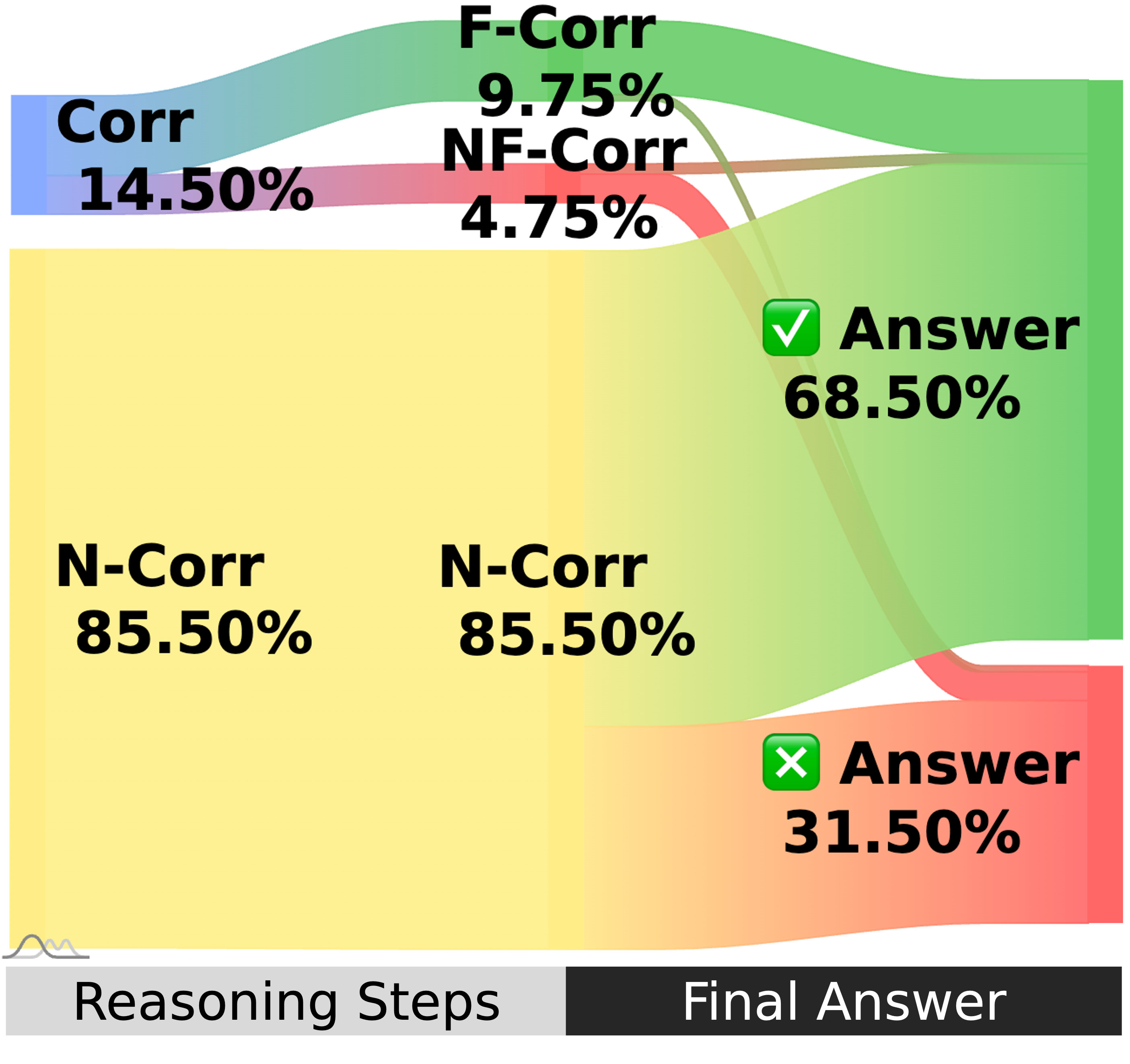}
      \caption{\textcolor{customorange}{Misinformed}}
      \label{fig:sankey_left}
  \end{subfigure}
  \begin{subfigure}[t]{0.235\textwidth}
      \centering
      \includegraphics[width=\textwidth]{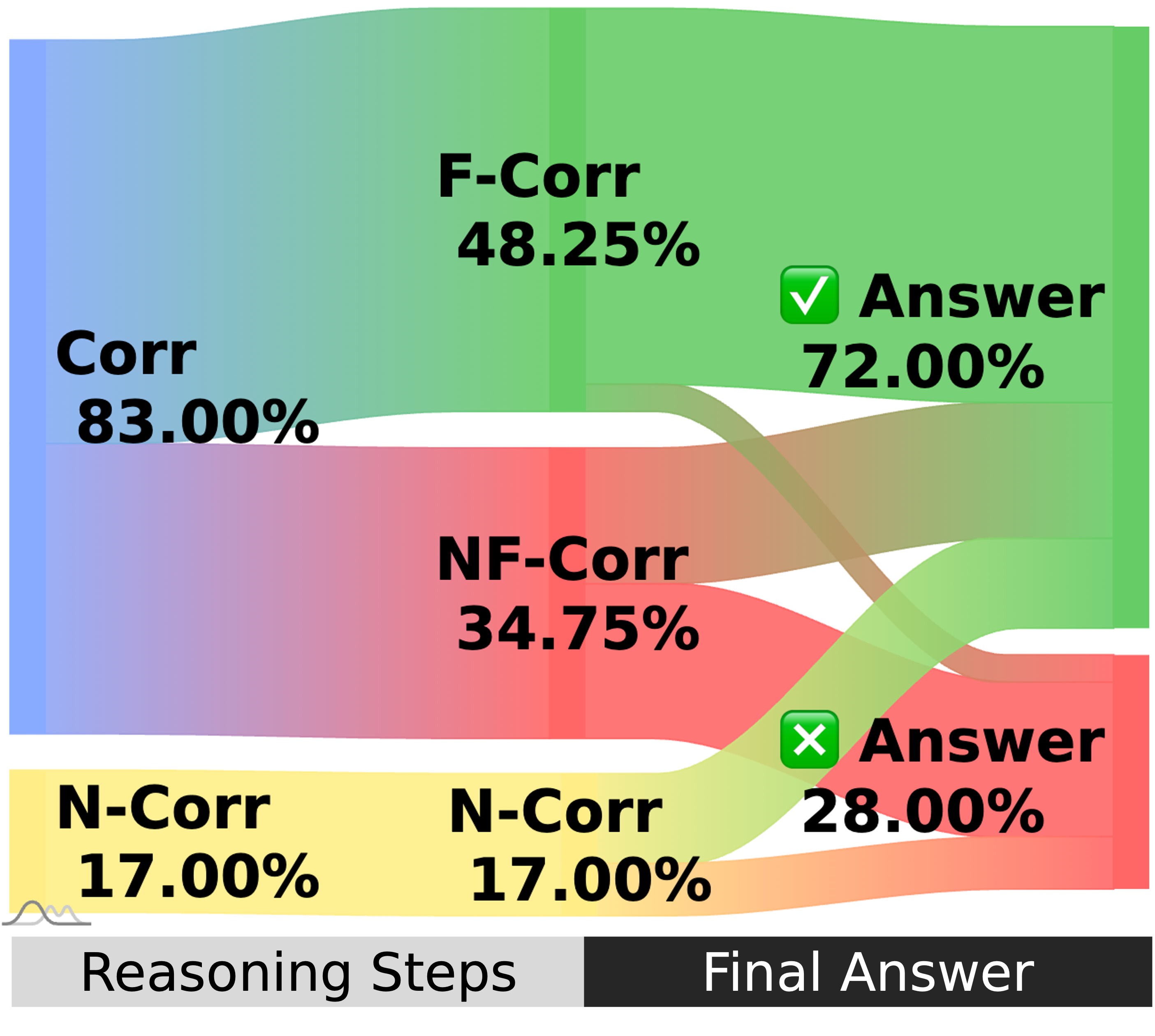}
      \caption{\textcolor{customgreen}{\Prompting}}
      \label{fig:sankey_right}
  \end{subfigure}
  \caption{Sankey diagrams from misinformation-correction behavior to K-Acc of outcome in (a) the \textcolor{customorange}{misinformed} setting and (b) the misinformed setting with the \textcolor{customgreen}{\prompting} instruction. The diagrams trace:
  (i) correction existence (\texttt{Corr}, \texttt{N-Corr}); 
  (ii) correction factuality (\texttt{F-Corr}, \texttt{NF-Corr}); and 
  (iii) final answer correctness.}
  \label{fig:sankey}
\end{figure}

\subsection{Default Behaviors Under Misinformation}
\label{sec:main:robustness:default}
Firstly, we quantify LLMs' default behavior under misinformation (denoted as \textcolor{customorange}{misinformed}). 
We evaluate their final answers and reasoning behaviors. 
As a comparison, we design \textcolor{customgreen}{\following} which instructs LLMs to always follow all given information, regardless of whether it is factual or not.


\SmallHeading{Final Answers}
As shown in \Table~\ref{tab:main_results}, by default, misinformation reduces \accuracy by 12.64\% -- 65.70\% (compared to \original \accuracy) on instruction models.
Meanwhile, with instructions to follow misinformation, \accuracy drops by 15.85\% -- 64.68\%. 
The great similarity in performance suggests that LLMs' default behavior to misinformation is more likely to follow it, though they have all the corresponding internal knowledge (by the definition of \accuracy).
More results are in \Appendix~\ref{sec:appendix:inst_follow}.

\begin{figure}[t!]
\centering
  \begin{subfigure}[t]{0.235\textwidth}
      \centering
      \includegraphics[width=\textwidth]{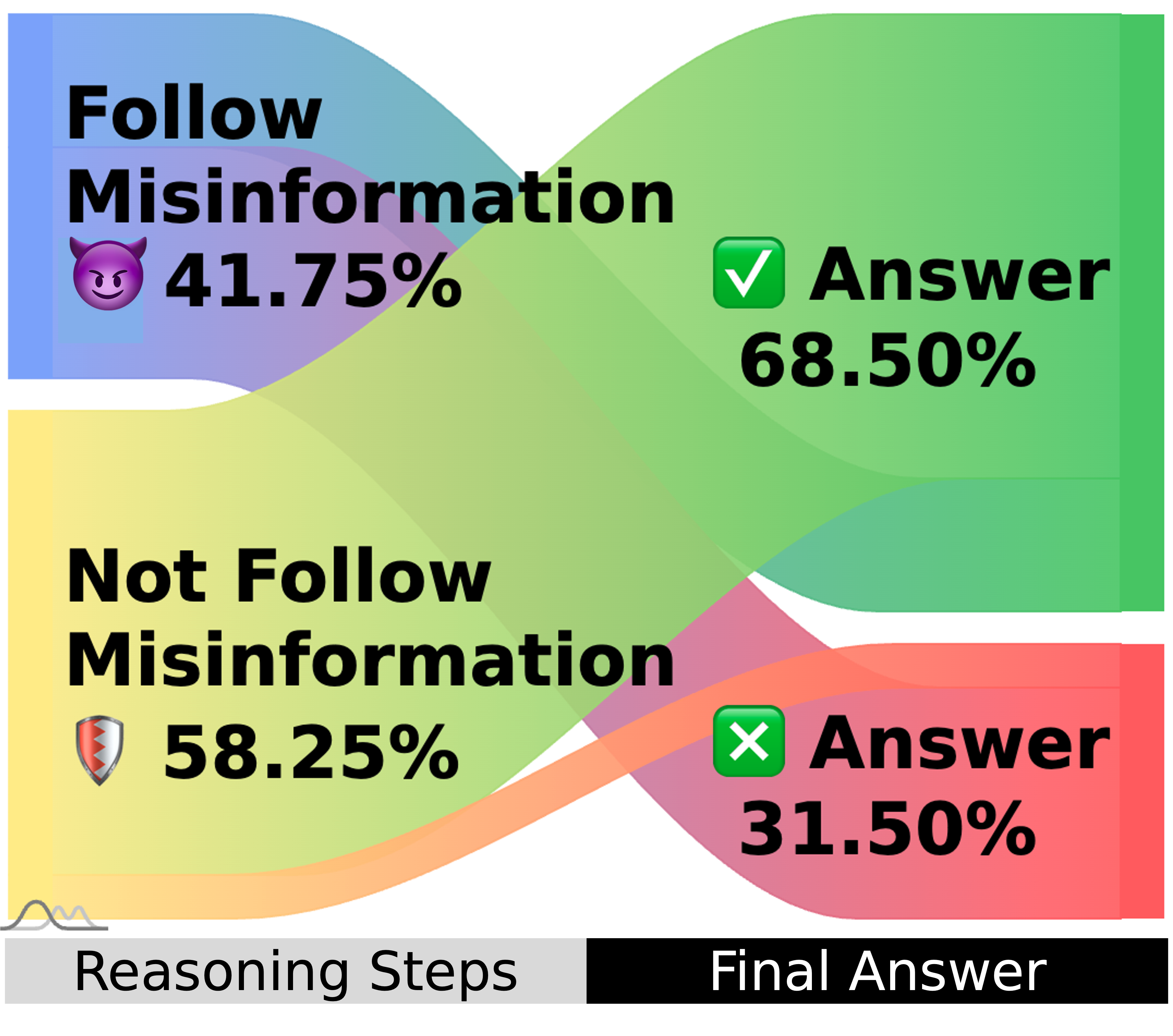}
      \caption{\textcolor{customorange}{Misinformed}}
      \label{fig:sankey_perturbed_left}
  \end{subfigure}
  \begin{subfigure}[t]{0.235\textwidth}
      \centering
      \includegraphics[width=\textwidth]{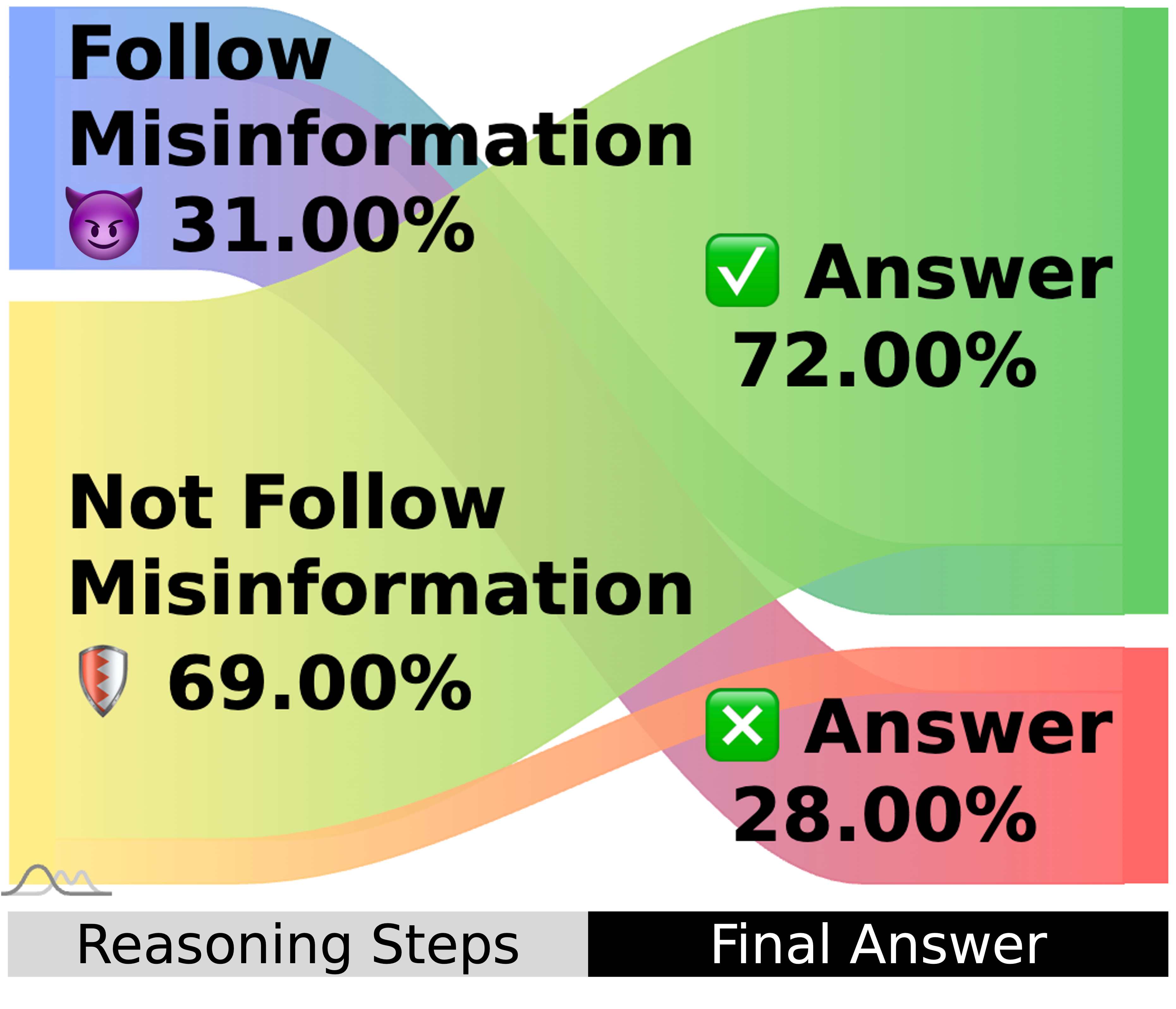}
      \caption{\textcolor{customgreen}{\prompting}}
      \label{fig:sankey_perturbed_right}
  \end{subfigure}
  \caption{Sankey diagrams from misinformation-following behavior in reasoning process to K-Acc of final answer in (\textit{a}) the \textcolor{customorange}{misinformed} setting and (\textit{b}) the misinformed setting with the \textcolor{customgreen}{\prompting} instruction.}
  \label{fig:sankey_perturbed}
\end{figure}

\SmallHeading{Reasoning Behaviors}
To understand why misinformation leads to more incorrect final answers, we also inspect the intermediate steps (\Section~\ref{sec:main:setup:analysis}). 
By default, LLMs correct misinformation in only 14.50\% of cases, with 9.75\% being factual corrections (\Figure~\ref{fig:sankey_left}). They follow 41.75\% of the misinformation, which contributes to incorrect reasoning in 63.47\% of cases (\Figure~\ref{fig:sankey_perturbed_left}).
As a comparison, \Figure~\ref{fig:sankey_follow} shows the behavior of \textcolor{customgreen}{\following}. The misinformation-following rate is 37.75\%, comparable to the rate of default behavior. 
This indicates that LLMs, by default, treat misinformation as an instruction to follow rather than something to correct, thereby lowering their accuracy. Thus, considering LLMs' instruction-following design \cite{ouyang2022training}, LLMs need explicit instructions to correct misinformation.

%% file: section/main/instruction.tex
\begin{figure*}[t!]
  \centering

  \begin{minipage}[t]{0.6\textwidth}
    \centering
    \raisebox{10pt}{
      \adjustbox{valign=t, width=\linewidth, keepaspectratio}{%
        \includegraphics{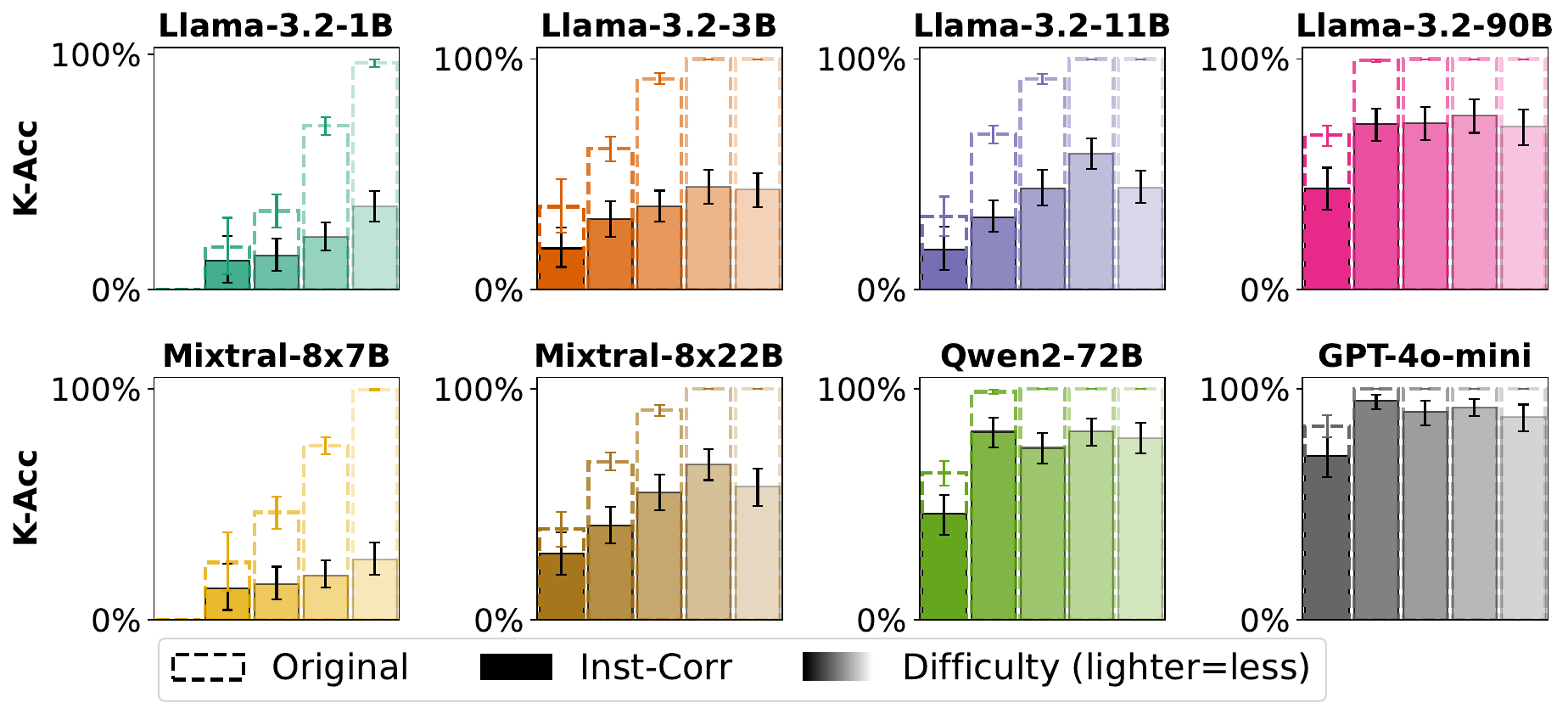}%
      }%
    }%
  \end{minipage}\hfill%
  \begin{minipage}[t]{0.4\textwidth}
    \adjustbox{valign=t, width=\columnwidth}{%
      \strut
      \resizebox{\columnwidth}{!}{%
        \renewcommand{\arraystretch}{1.15}
        \setlength{\tabcolsep}{5pt}
        \input{section/appendix/big_table/pearson_correlation}
        
      }
    }
  \end{minipage}

  \caption{
    \textit{Left}: \Accuracy (with 95\% confidence intervals) across LLMs in the \textcolor{customblue}{\original} setting (dashed-border bars) and the misinformed setting with \textcolor{customgreen}{\prompting} instructions (solid bars), stratified by question difficulty (lighter shades indicate easier questions).
    \textit{Right}: Pearson correlation coefficients and corresponding p-values between question difficulty (1-5, higher=easier) and the relative \accuracy decrease from the \textcolor{customblue}{\original} setting to the misinformed setting with \textcolor{customgreen}{\prompting} instructions.
  }
  \label{fig:bar}
\end{figure*}

\subsection{Instruct to Correct Misinformation}
\label{sec:main:correction_methods:prompting}


In this section, we explicitly instruct models to correct any misinformation in LLM inputs (denoted as \textcolor{customgreen}{\prompting}).
Since models have correct internal knowledge on our evaluated questions (by the definition of \accuracy), they should potentially be able to correct misinformation when explicitly instructed. 

\SmallHeading{Final Answers}
As shown in \Table~\ref{tab:main_results}, explicit \prompting instructions improve misinformed performance for 5 models, but still lag behind \original \accuracy by 10.02\% to 72.20\%. There is even a backfire effect in smaller models like Llama-3.2-1B, where \prompting \accuracy is lower than \misinformed. 
We also compare misinformed with original \accuracy on questions with different difficulty gradations (\Figure~\ref{fig:bar}).\footnote{We sample five final answers $a^{(k)}$ {\small{$(k=1, \cdots, 5)$}} for each $q$ and measure the percentage of correct $a^{(k)}$. We then sort questions by the percentage and evenly divide them into five bins as five difficulty gradations. Note that the divisions are relative to each model, as models' knowledge differs.} Even with explicit \prompting instructions, misinformed \accuracy is lower than the original one across all difficulty gradations of all models. We also measure the overall Pearson correlation between the difficulty gradation\footnote{We assign integers from 1 to 5 to represent the question difficulty gradation, where larger values indicate less difficulty.} and the relative decrease from \original to \misinformed K-Acc, and observe no significant correlation across all models (all p-values > 0.05). This suggests that LLMs' vulnerability to misinformation persists across all models and question difficulty levels, even when they are instructed to use their internal knowledge to correct misinformation.


\SmallHeading{Reasoning Behaviors}
Even with explicit \prompting instructions to correct misinformation on questions where LLMs possess correct internal knowledge,
only 48.25\% is factually corrected (\Figure~\ref{fig:sankey_right}). For the small model (Llama-3.2-1B), nonfactual corrections become more frequent (\Figure~\ref{fig:sankey_1b}): \prompting improves the overall correction frequency from 2.25\% to 51.50\% but increases nonfactual corrections significantly (from 1.00\% to 47.75\%), ultimately decreasing \accuracy from 46.75\% to 39.25\%.
We also observe that LLMs still follow misinformation 31.00\% of time (\Figure~\ref{fig:sankey_perturbed_right}). Thus, we conclude that LLMs struggle to correct misinformation via explicit instructions, revealing their limited steerability towards misinformation correction.

\begin{figure*}[t!]
  \begin{center}
    \includegraphics[width=0.99\textwidth]{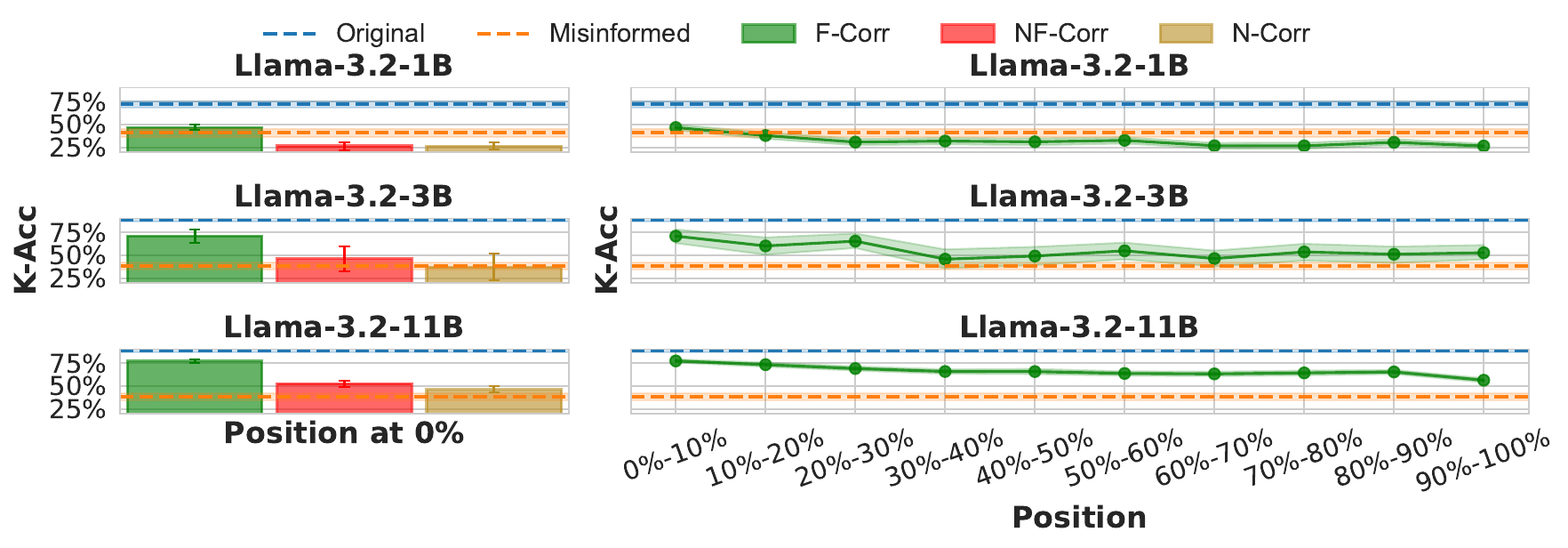}
  \end{center}
  \caption{\textit{Left}: \Accuracy across  three correction behaviors, \texttt{F-Corr}, \texttt{NF-Corr}, and \texttt{N-Corr}, at the first reasoning step (0\% position). The \textcolor{customorange}{misinformed} dashed line indicates the baseline performance, while the \textcolor{customblue}{original} dashed line indicates the upper-bound of recovered performance (from the original setting with no misinformation). 
  \textit{Right}: \Accuracy of factual corrections (\texttt{F-Corr}) across different positions in the CoT process under the \misinformed setting. The x-axis denotes the position at which the correction occurs. Shaded areas represent 95\% confidence intervals.
  }
\label{fig:pathway}
\end{figure*}


\subsection{Impact on Thinking Models}
\label{sec:main:thinking_models}

Recently, reasoning LLMs with thinking \cite{jaech2024openai, guo2025deepseek} have empowered more advanced math reasoning capabilities.
In \Section~\ref{sec:main:robustness:default} and \Section~\ref{sec:main:correction_methods:prompting}, we mainly focus on instruction models, which as the backbone for thinking models. 
In this section, we experiment with more thinking models to assess whether our conclusions of misinformation propagation still persist.

As shown in \Table~\ref{tab:thinking_results}, our observations still hold for thinking models. 
First, by default, misinformation reduces K-Acc by 6.62\% to 62.78\%. Even with instructions to follow misinformation, the performance drop is still similar (11.18\% -- 62.92\%), showing that thinking models still follow misinformation as instructions by default.
Besides, the performance gap between the original \accuracy and misinformed \accuracy with explicit instructions (4.30\% -- 19.97\%) to correct misinformation indicates that thinking models still fail to correct misinformation on all questions where they possess internal knowledge, even when they are instructed to correct misinformation. Overall, the vulnerability and limited steerability exhibit in both instruction and thinking models, suggesting they originate in the instruction-tuning stage and cannot be fully resolved through additional reasoning post-training. Hence, following on, our analysis focuses more on instruction models.

%% file: section/appendix/big_table/pearson_correlation.tex


\begin{tabular}{>{\bfseries}p{2.0cm}|>{\raggedleft\arraybackslash}p{2.5cm}>{\raggedleft\arraybackslash}p{2.5cm}>{\raggedleft\arraybackslash}p{2.5cm}>{\raggedleft\arraybackslash}p{2.5cm}>{\raggedleft\arraybackslash}p{2.5cm}}
\rowcolor{gray!25}
\toprule
Pearson Corr. & \textit{\textbf{All Models}} & Llama-3.2-1B & Llama-3.2-3B \\
\midrule
Coefficients & 0.21 & 0.84 & 0.65\\
P-Values      & 0.21 & 0.16 & 0.24\\
\midrule
\rowcolor{gray!25}
& Llama-3.2-11B & Llama-3.2-90B & Qwen-2-72B\\
Coefficients & 0.24 & -0.59 & -0.43\\
P-Values      & 0.69 & 0.30 & 0.47\\
\midrule
\rowcolor{gray!25}
& Mixtral-8×7B & Mixtral-8×22B & GPT-4o-mini\\
\midrule
Coefficients & 0.88 & 0.57 & -0.16 \\
P-Values      & 0.12 & 0.31 & 0.79 \\
\bottomrule
\end{tabular}%


%% file: section/main/potential_pathways.tex
\section{Mitigating Propagation via Correction}
\label{sec:main:controlled}


To reliably steer models to mitigate misinformation propagation via correction, we
(\textit{i}) investigate what factors make corrections more effective (\Section~\ref{sec:main:controlled:factors}) and
(\textit{ii}) whether fine-tuning with effective corrections mitigates propagation and improves reasoning factuality (\Section~\ref{sec:main:controlled:finetuning}).

\takeawaysc{
(\textit{i}) Early factual corrections are most effective at mitigating misinformation propagation.
(\textit{ii}) Fine-tuning models for early corrections significantly improves reasoning performance under misinformation, although still not recovering the original performance.
}

\subsection{Factors of Effective Correction} \label{sec:main:controlled:factors}

Since LLMs correct misinformation with different behaviors (\Section~\ref{sec:main:robustness:default}) and positions (\Figure~\ref{fig:position_distribution}), we explore how these factors affect correction effectiveness.
In this section, we conduct a controlled study by enforcing correction behaviors (\ie{} \texttt{N-Corr}, \texttt{F-Corr}, and \texttt{NF-Corr}) at different positions of the reasoning steps using Llama 3.2 models.
Specifically, we change LLM's output of a certain reasoning step to exhibit different correction behaviors (\Section~\ref{sec:main:setup:analysis}).
For instance, we set the content of a step as the correction to the truthful equation to enforce \texttt{F-Corr}. Detail prompt designs are at \Appendix~\ref{sec:appendix:prompt_design:self_correction_cot_syn}.

\SmallHeading{Effects of Correction Behaviors}
First, to analyze the impact of different correct behaviors, we enforce one of the three distinct behaviors at the initial reasoning step, $c_1$: factual correction (\texttt{F-Corr}), non-factual correction (\texttt{NF-Corr}), or no correction (\texttt{N-Corr}).
In \Figure~\ref{fig:pathway} (Left), we observe how different correction behaviors improve or decrease misinformed \accuracy and compare their deviations from original \accuracy, representing the upper bound achievable by corrections.
We show that factually correcting misinformation improves \misinformed \accuracy by 14.35\%$_{\text{(1B)}}$, 84.03\%$_{\text{(3B)}}$, and 101.58\%$_{\text{(11B)}}$, though the performance still lags behind the \original 
by 35.05\%$_{\text{(1B)}}$, 19.90\%$_{\text{(3B)}}$, and 12.70\%$_{\text{(11B)}}$.
In contrast, nonfactual or no corrections reduce the \misinformed \accuracy by 35.06\%$_{\text{(1B)}}$ or yield only 7.95\%$_{\text{(3B)}}$ and 28.88\%$_{\text{(11B)}}$ gains.
We argue that factual corrections are necessary for mitigating misinformation propagation across all model sizes, whereas nonfactual or absent corrections offer limited benefits and can be counterproductive, especially for smaller models.
\begin{figure*}[t!]
  \centering
  \begin{subfigure}[t]{0.245\textwidth}
      \centering
      \includegraphics[width=\textwidth]{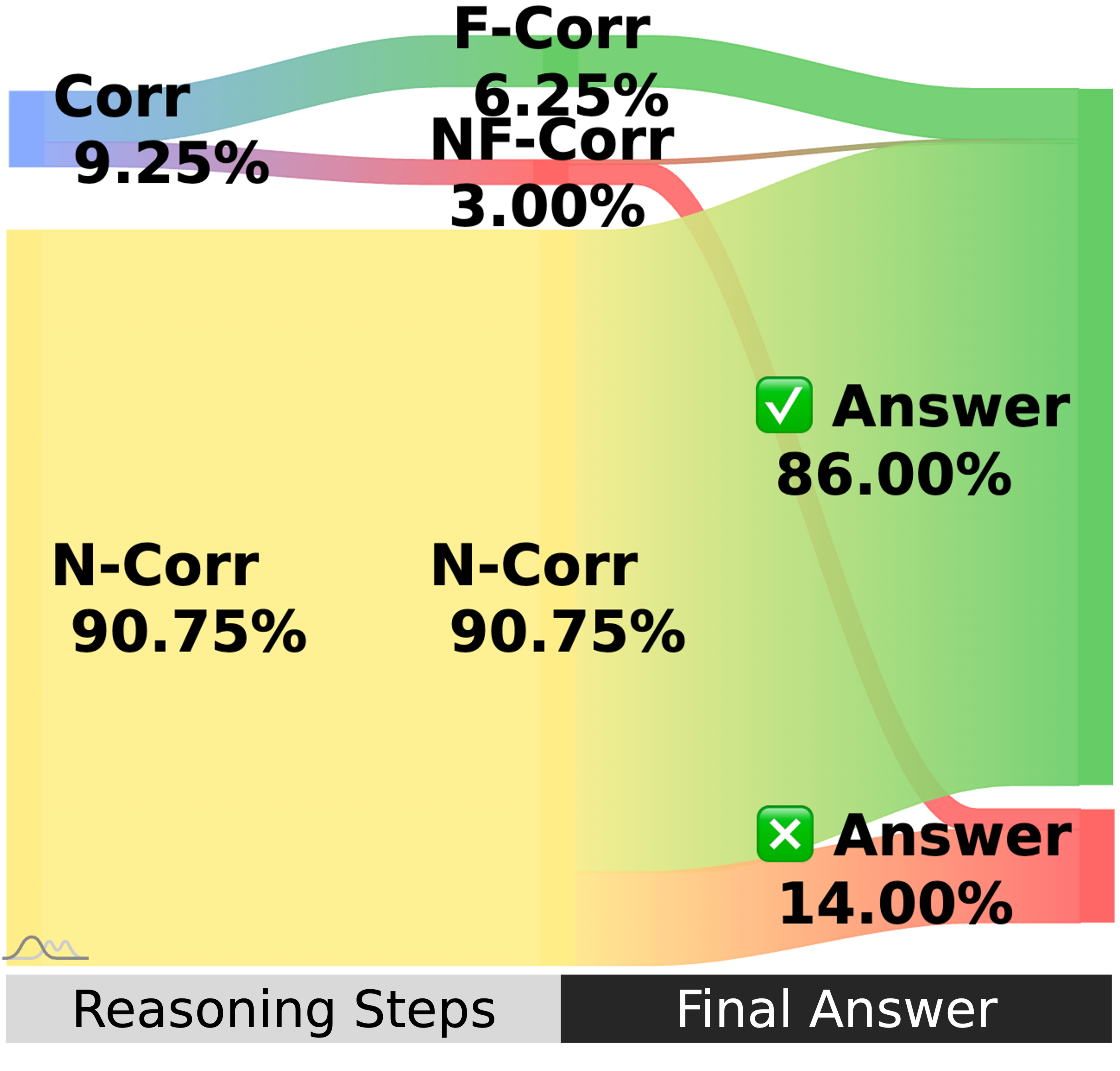}
      \caption{Base Model}
      \label{fig:sankey_ft_base}
  \end{subfigure}
  \begin{subfigure}[t]{0.245\textwidth}
      \centering
      \includegraphics[width=\textwidth]{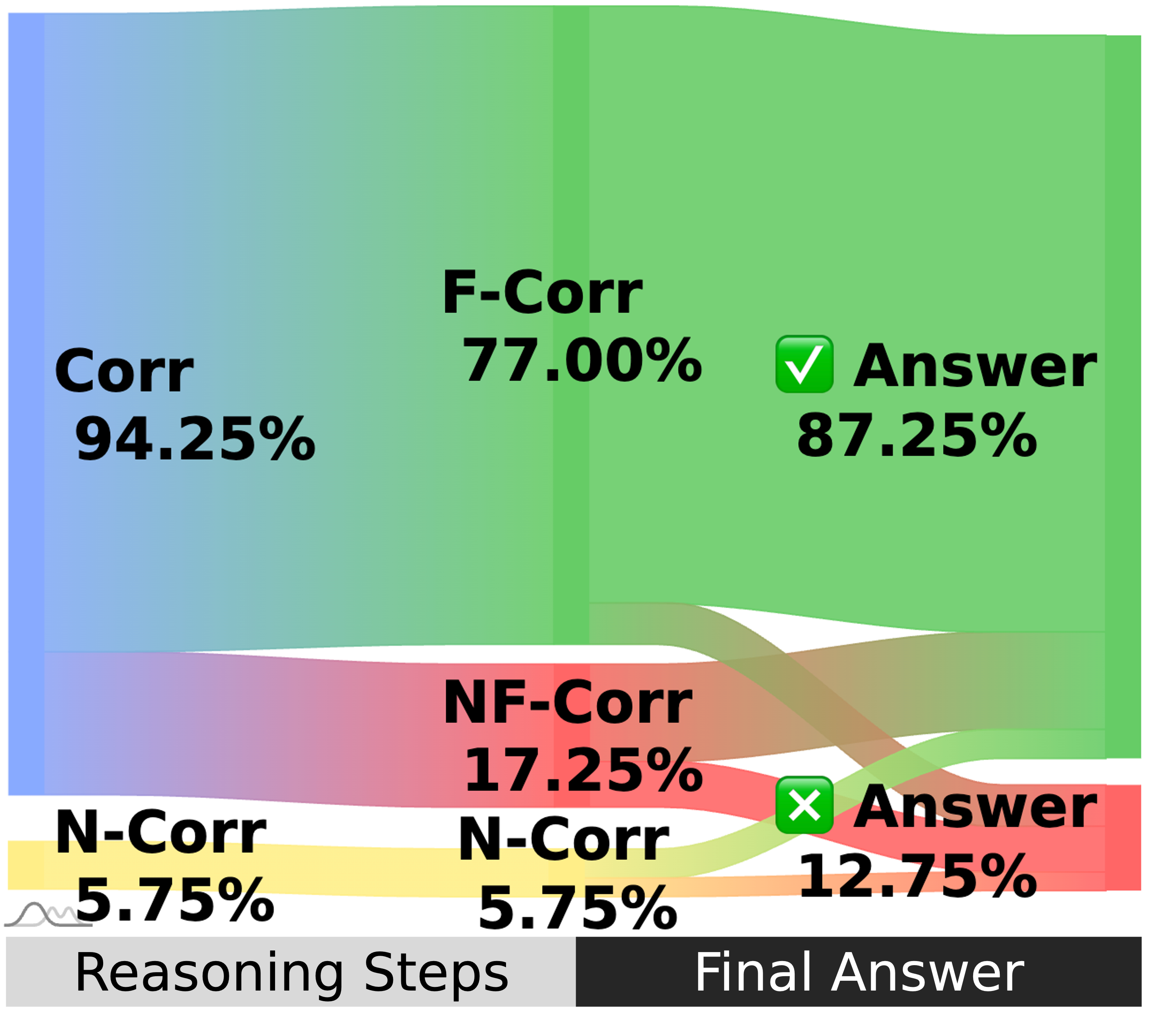}
      \caption{\textcolor{customgreen}{\Prompting}}
      \label{fig:sankey_ft_inst}
  \end{subfigure}
  \begin{subfigure}[t]{0.245\textwidth}
      \centering
      \includegraphics[width=\textwidth]{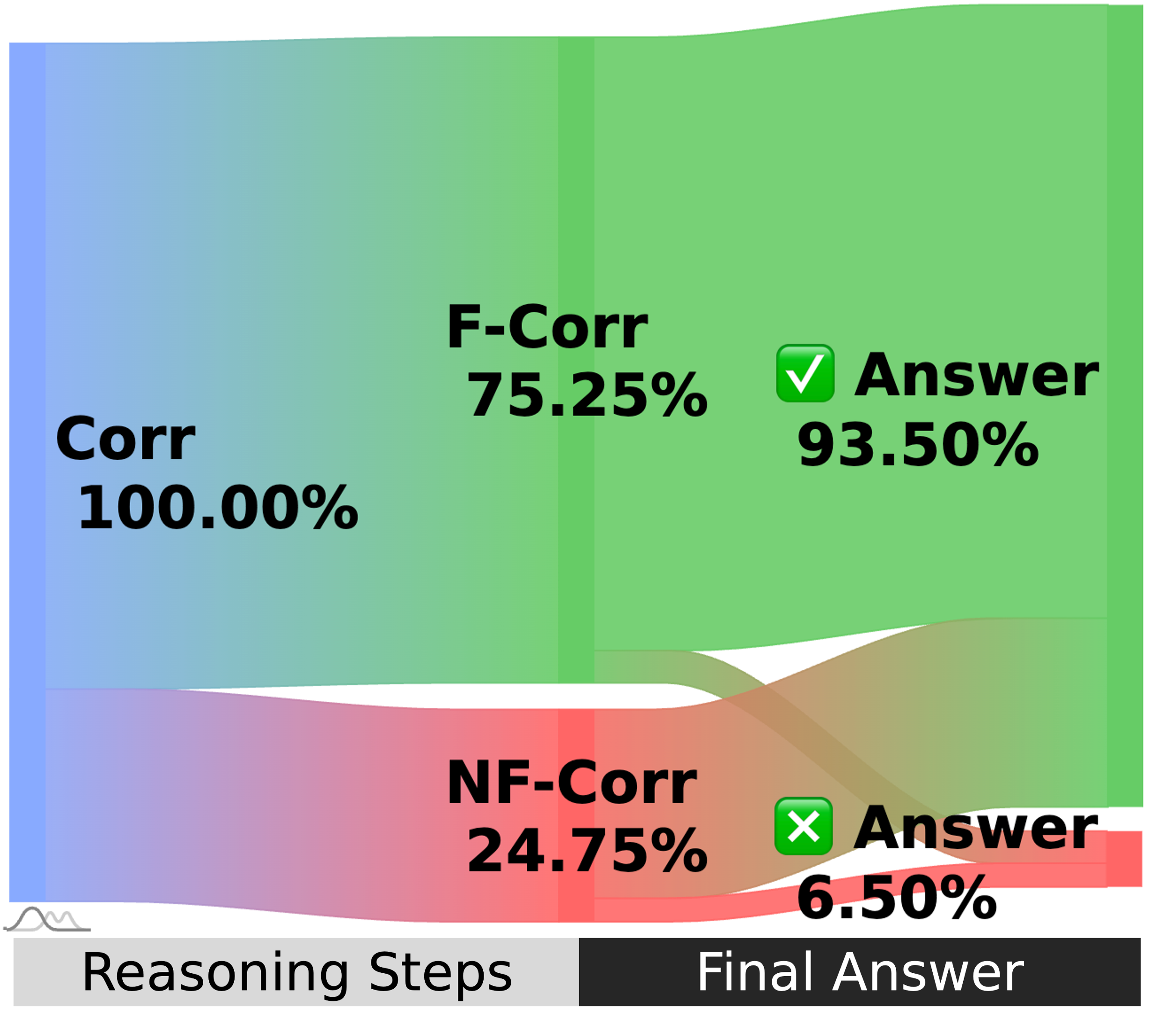}
      \caption{\textcolor{violet}{\Finetuning}}
      \label{fig:sankey_ft_ft}
  \end{subfigure}
  \begin{subfigure}[t]{0.245\textwidth}
      \centering
      \includegraphics[width=\textwidth]{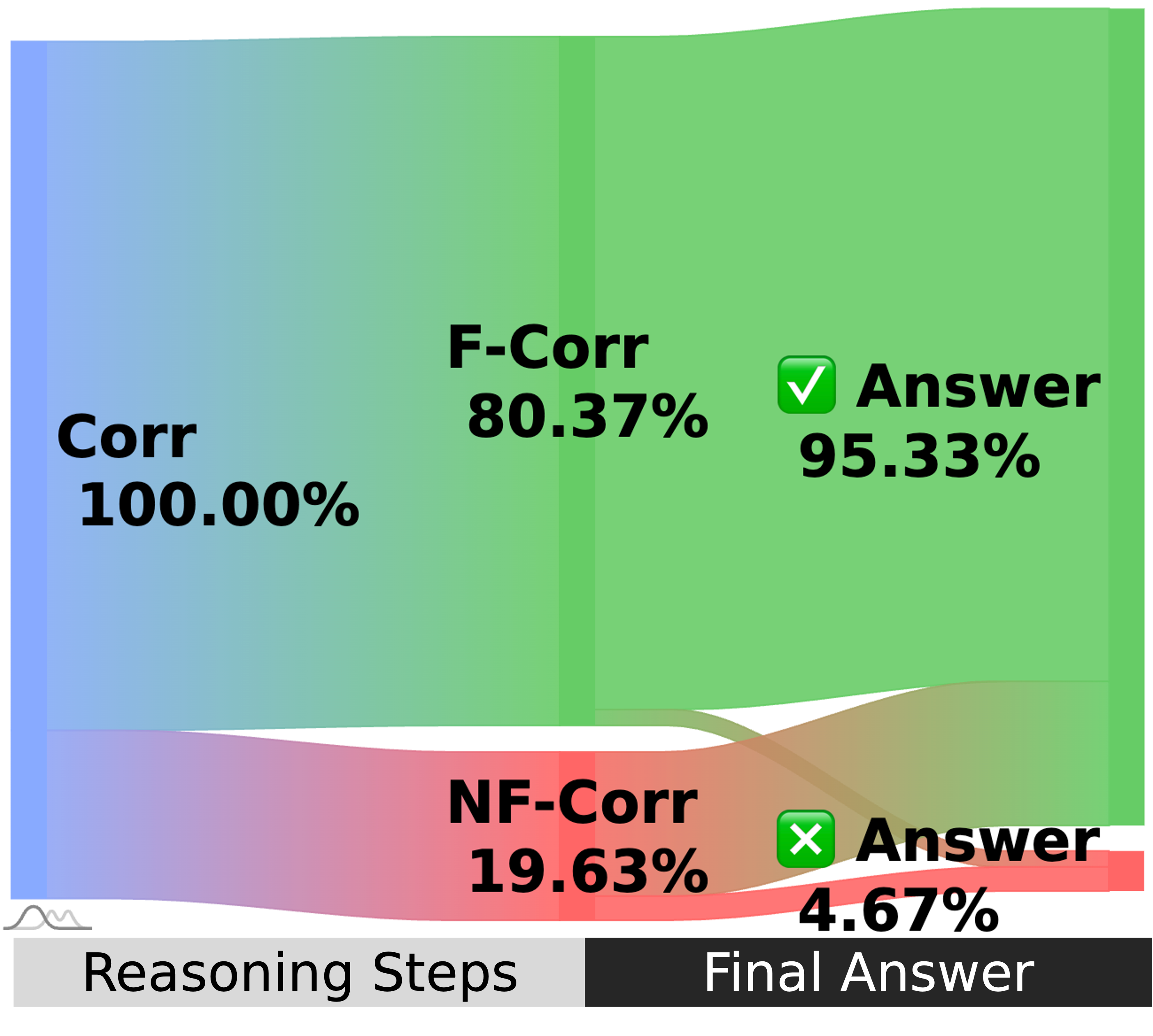}
      \caption{\textcolor{customgreen}{\Prompting} \textcolor{violet}{+ \Finetuning}}
      \label{fig:sankey_ft_inst_ft}
  \end{subfigure}
  \caption{Sankey diagram illustrating the flow of GPT-4o-mini’s reasoning from misinformation correction, through factual correction, to \Accuracy. We compare four setups: the base GPT-4o-mini model, the model with only misinformation-correction instructions (\textcolor{customgreen}{\prompting}), the fine-tuned models without instructions (\textcolor{violet}{\finetuning}), and the fine-tuned model with instructions (\textcolor{customgreen}{\prompting} + \textcolor{violet}{\finetuning}). All evaluations are conducted on instances where the base model demonstrates correct knowledge in the original setting.}
  \label{fig:sankey_ft}
\end{figure*}
\SmallHeading{Effects of Correction Positions}
Next, we vary the position of factual corrections during reasoning steps, by injecting \texttt{F-Corr} at different $c_i$.
In \Figure~\ref{fig:pathway} (Right), we plot resulting \accuracy trends as correction position changes, alongside comparisons with original and misinformed \accuracy.
We show that corrections at the beginning of reasoning steps (<10\%) are more effective in improving final answer accuracy, achieving the highest \accuracy. Performance declines as corrections are applied at later steps. For the 1B model, \accuracy with factual correction could be even worse than the \misinformed setting simply after 10\% of steps. These results suggest that earlier corrections are more effective for misinformation mitigation. Delayed corrections may become ineffective or even detrimental, especially for smaller models, likely due to the accumulation of erroneous information early in the reasoning process that becomes difficult to correct later. 




\subsection{Fine-tuning with Effective Correction}
\label{sec:main:controlled:finetuning}

Findings from \Section~\ref{sec:main:controlled:factors} indicate that early factual corrections are effective. Given that most instruction models are not fine-tuned on correction-specific data \cite{kamoi2024can}, we fine-tune LLMs to perform early factual corrections to analyze whether fine-tuning improves reasoning factuality under misinformation. Each fine-tuning data point includes a factual correction in the first reasoning step. Details on fine-tuning prompt design are provided in \Appendix~\ref{sec:appendix:prompt_design:mitigation:fine_tune}, while data collection and training procedures are in \Appendix~\ref{sec:appendix:model_setup}.

\SmallHeading{Significant Performance Improvements}
\Table~\ref{tab:finetune} compares performance in original and misinformed scenarios across four setups: 
the base GPT-4o-mini model (Base Model), 
the base model with explicit correction instructions (\textcolor{customgreen}{\Prompting}), the base model with fine-tuning only (\textcolor{violet}{\Finetuning}), and the base model with both (\textcolor{customgreen}{\Prompting} + \textcolor{violet}{\Finetuning}).
While explicit correction instructions (\textcolor{customgreen}{\Prompting}) only raise the misinformed \accuracy from 85.64\% to 88.21\%, fine-tuning (\textcolor{violet}{\Finetuning}) significantly boosts performance to 95.68\%, nearing the original performance (98.03\%). However, all intervention methods slightly reduce performance on original questions: \original \accuracy drops from 98.03\% to 96.13\% (\textcolor{customgreen}{\Prompting}), 95.31\% (\textcolor{violet}{\Finetuning}), and 95.71\% (\textcolor{customgreen}{\Prompting} + \textcolor{violet}{\Finetuning}). This indicates that fine-tuning for early factual corrections enhances the model's ability to correct misinformation and reason factually with a minor trade-off in \accuracy on original, non-misinformed questions.
Apart from the GPT-4o-mini model, we also fine-tune an open-source instruction model (Llama-3.2-3B) and a thinking model (DeepSeek-R1-Distilled-Qwen-2.5-1.5B). As shown in \Table~\ref{tab:apdx_finetune}, fine-tuning improves misinformed \accuracy from 38.41\% to 44.11\% for Llama-3.2-3B and 71.47\% to 75.64\% for DeepSeek-R1-Distilled-Qwen-2.5-1.5B with explicit instructions to correct misinformation.
This indicates that the effectiveness of fine-tuning in mitigating misinformation propagation still holds for advanced reasoning models.

\begin{table}[t!]
    \centering
    \resizebox{0.48\textwidth}{!}{
    \input{section/appendix/big_table/finetune}
    }
    \caption{\Accuracy (\%) with 95\% confidence intervals under the original and misinformed settings (GPT-4o-mini). We compare the same four setups as \Figure~\ref{fig:sankey_ft}. ``Gap'' indicates the relative accuracy drop from the original to the misinformed setting. All evaluations are conducted on instances where the base model demonstrates correct knowledge in the original setting.}
    \label{tab:finetune}
\end{table}

\SmallHeading{Boost of Correction Behaviors}
\Figure~\ref{fig:sankey_ft} illustrates the changes in correction behavior under the misinformed setting across the different setups. Explicit instructions (\textcolor{customgreen}{\Prompting}) increase the correction frequency from 9.25\% (Base Model) to 94.25\% (\Figure~\ref{fig:sankey_ft_base} vs. \Figure~\ref{fig:sankey_ft_inst}). Fine-tuning (\textcolor{violet}{\Finetuning}), either alone or with instructions, achieves a 100\% correction frequency (\Figure~\ref{fig:sankey_ft_ft} and \Figure~\ref{fig:sankey_ft_inst_ft}). The combination of fine-tuning and explicit instructions (\textcolor{customgreen}{\Prompting} + \textcolor{violet}{\Finetuning}) yields the highest factual correction ratio (80.37\%) and improves the overall misinformed \accuracy to 95.33\%. This represents a significant improvement compared to the base model under misinformation.

\begin{table}[t!]
    \centering
    \resizebox{0.48\textwidth}{!}{
    \input{section/appendix/big_table/apdx_finetune}
    }
    \caption{Misinformed \Accuracy (\%) with 95\% confidence intervals with instructions to correct misinformation before and after fine-tuning. We test one close-source instruction model (GPT-4o-mini), one open-source instruction model (Llama-3.2-3B), and one open-source thinking model (DeepSeek-R1-Distilled-Qwen-2.5-1.5B). ``$\uparrow$'' represents relative increase after fine-tuning. All evaluations are conducted on instances where the base model demonstrates correct knowledge in the original setting.}
    \label{tab:apdx_finetune}
\end{table}

%% file: section/appendix/big_table/finetune.tex
\begin{tabular}{p{2.3cm}|p{2.3cm}p{2.3cm}p{1.8cm}}
\toprule
& \textcolor{customblue}{\Original} & \textcolor{customorange}{\Misinformed} & \textcolor{customgreen}{Gap} \\ \midgrayline
Base Model & 98.03 \tiny{[97.31, 98.62]} & 85.64 \tiny{[82.75, 88.38]} & $\downarrow$12.64\% \\
\midgrayline
+ \textcolor{customgreen}{\prompting} & 96.13 \tiny{[94.83, 97.31]} & 88.21 \tiny{[85.74, 90.51]} & $\downarrow$ 8.24\% \\
\midgrayline
+ \textcolor{violet}{\finetuning} & 95.31 \tiny{[93.83, 96.71]} & 95.11 \tiny{[93.38, 96.53]} & $\downarrow$ 2.09\% \\
\midgrayline
\makecell[l]{+ \textcolor{customgreen}{\prompting}\\+ \textcolor{violet}{\finetuning}} & 95.71 \tiny{[94.33, 96.94]} & 95.68 \tiny{[94.24, 96.96]} & $\downarrow$ 0.03\% \\
\bottomrule
\end{tabular}

%% file: section/appendix/big_table/apdx_finetune.tex
\begin{tabular}{p{2.1cm}|p{2.4cm}p{2.4cm}p{2.4cm}}
\toprule
Base Model & GPT-4o-mini & Llama-3.2-3B & DeepSeek-R1-Distilled-Qwen-2.5-1.5B \\ \midgrayline
\makecell[l]{\textcolor{customorange}{Misinformed}\\+\textcolor{customgreen}{\prompting}} & 88.21 \tiny{[85.74, 88.38]} & 38.41 \tiny{[34.79, 41.91]} & 71.47 \tiny{[68.36, 74.76]} \\
\midgrayline
\makecell[l]{\textcolor{customorange}{Misinformed}\\+\textcolor{customgreen}{\prompting}\\+ \textcolor{violet}{\finetuning}} & 95.68 \SPSB{\tiny{$\uparrow$ 8.47\%}}{\tiny{[94.24, 96.96]}} & 44.11 \SPSB{\tiny{$\uparrow$ 14.84\%}}{\tiny{[42.05, 50.09]}} & 75.64 \SPSB{\tiny{$\uparrow$ 5.83\%}}{\tiny{[72.73, 78.74]}} \\
\bottomrule
\end{tabular}

%% file: section/main/conclusion.tex
\section{Conclusion}
We systematically study the impact of misinformation on LLM reasoning and mitigation effectiveness. Misinformation reduces LLM reasoning accuracy and LLMs struggle to factually correct it, even with correct internal knowledge and explicit instructions. This reveals LLMs' vulnerability to misinformation and limited steerability to mitigate its propagation. We show two remedies: (\textit{i}) early factual corrections in reasoning, and (\textit{ii}) fine-tuning for these corrections, which sharply mitigate misinformation propagation and restore factuality, offering valuable insights for handling misinformation.

%% file: section/main/acknowledgements.tex
\section*{Acknowledgments}
We gratefully acknowledge the financial and IT support provided by EPFL, which allows us to use the server and computing resources. We also thank TogetherAI for providing API credits.

%% file: section/main/limitation.tex
\section*{Limitations}
Our work explores misinformation propagation, including its impact and mitigation.
However, our study has limitations on the scope of reasoning tasks, methods, and models.
We only study math reasoning tasks and focus on text-based interaction, which have ground-truth answers for reliable evaluation and are simpler to synthesize misinformation, and concentrate on CoT reasoning for easier final-answer extraction and manipulation, though other reasoning methods exist, for example, Tree-of-Thought (ToT)~\cite{yao2023tree} and scratchpad~\cite{nye2021show}.
Besides, we only deploy open source models with fewer than 15 B parameters locally due to computational overheads. Our curated test data could also be expanded to a larger scale.
Motivated by these shortcomings, future work could explore the effect of misinformation on more reasoning tasks, methods, and more diverse LLMs.

As discussed in \Section~\ref{sec:main:related}, while our work emphasizes correcting misinformation that conflicts with models' internal knowledge with explicit instructions, follow such information is intended in certain scenarios, like counterfactual settings (e.g., an octal number system where 7+1=10 holds true). Our initial studies on instructing models to follow misinformation (\Appendix~\ref{sec:appendix:inst_follow}) show that their \accuracy remains well above zero (\Table~\ref{tab:follow_results}), with models following the misinformation only 37.75\% of the time (\Figure~\ref{fig:sankey_follow}). This suggests models struggle to consistently follow instructions that contradict their internal knowledge and face difficulty with user-provided counterfactual information. Developing more comprehensive settings to study counterfactual instruction-following is an important direction for future work.

%% file: section/main/ethical_consideration.tex
\section*{Ethical Considerations}
Our work shows that misinformation often leads to reasoning errors in LLMs, providing malicious users with new ways to attack models by injecting errors into the input. 
However, as mentioned in the introduction, \citet{kumar2023math} and \citet{xu2024ai} have already noted that users may unknowingly include errors in their queries in mathematical contexts. 
Hence, rather than focusing solely on the risk of malicious misinformation attacks, it is even more important to understand how LLMs process and correct unintended misinformation. This ensures that LLMs not only avoid being misled into incorrect outputs, which could negatively impact users, but also help users recognize and correct their own mistakes. 
By presenting the full results of our study, we aim to raise awareness of misinformation in the LLM community and improve the reliability of LLM reasoning.

%% file: appendix.tex
\input{section/appendix/user_misinformation}

\input{section/appendix/prompt_design}

\input{section/appendix/evaluation_framework}

\input{section/appendix/model_setup}

\input{section/appendix/inst_follow}

\input{section/appendix/checklist}



%% file: section/appendix/user_misinformation.tex
\section{Table of Contents} \label{sec:appendix:toc}

This table of contents outlines the structure of the appendix to facilitate easier navigation.

\begin{itemize}
    \item \textbf{Testing Data Collection} (\Appendix~\ref{sec:appendix:test_data_collection})---This section details our methodology for selecting and processing data from canonical math datasets (\Appendix~\ref{sec:appendix:data_processing}). It also explains how we simulate misinformation from the gathered math questions (\Appendix~\ref{sec:appendix:user_misinformation}).
    \item \textbf{Prompt Design} (\Appendix~\ref{sec:appendix:experiment_design})---This section provides details on how we design prompts to assess the impact of misinformation (\Appendix~\ref{sec:appendix:prompt_design}) and to mitigate its propagation (\Appendix~\ref{sec:appendix:prompt_design:mitigation}).
    \item \textbf{Evaluation Framework} (\Appendix~\ref{sec:appendix:cot_evaluation})---This section details our framework for evaluating final answers (\Appendix~\ref{sec:appendix:cot_evaluation:accuracy}) and reasoning steps (\Appendix~\ref{sec:appendix:cot_evaluation:step}).
    \item \textbf{Model Setup} (\Appendix~\ref{sec:appendix:model_setup})---This section provides details on the model setup for both fine-tuning and inference.
    \item \textbf{Additional Results} (\Appendix~\ref{sec:appendix:additional_results})---This section presents additional experimental results for scenarios where LLMs are instructed to follow misinformation (\Appendix~\ref{sec:appendix:inst_follow}) or correct misinformation (\Appendix~\ref{sec:appendix:inst_corr}).
    \item \textbf{Responsible NLP Research} (\Appendix~\ref{sec:appendix:checklist})---This section lists the main data artifacts, backbone models, and major packages used (\Appendix~\ref{sec:appendix:checklist:artifacts}). It also reports details about the annotators involved in assessing reasoning correctness and behaviors (\Appendix~\ref{sec:appendix:checklist:annotators}). Finally, it clarifies our use of AI in coding and writing (\Appendix~\ref{sec:appendix:checklist:ai}).
\end{itemize}

\section{Testing Data Collection} \label{sec:appendix:test_data_collection}

\input{section/appendix/data_processing}

\subsection{Misinformation Simulation}
\label{sec:appendix:user_misinformation}

\subsubsection{Truthful Equation Generation}

We use either an external LLM (\texttt{gpt-4-0613}) or the tested LLM itself to generate equations. Prompts include a \colorbox{blue!30}{system} and a \colorbox{yellow!30}{user} message, where the system message further includes an instruction and a demonstration:

\begin{mdframed}[backgroundcolor=gray!20,linewidth=0pt] 
    \begin{framedwithtag}{System}{blue!30}{blue!10}{-50mm}{2mm}{%
    You are given a question. Generate only LaTeX formulas for the question without ever answering the question or revealing the answer. Each formula should be wrapped between single dollar signs and separated by semicolons. The variables should be either from the question or wrapped in \$\textbackslash text\{...\}\$. 
    
    Example:
    
    Question: \{an example question\}
    
    Answer: \{ground-truth equations\}
    }
    \end{framedwithtag}
    \vspace{-10mm}
    \begin{framedwithtag}{User}{yellow!30}{yellow!10}{-30mm}{2mm}{%
    Question: \{a test question\}
    
    Answer:
    }
    \end{framedwithtag}
\end{mdframed}

\subsubsection{Heuristic Rule Generation}

We apply heuristic rules to simulate human-like erroneous equations, including numeric value modification, operator alteration, and operand swapping. Specifically, we use \texttt{gpt-4o-mini-2024-07-18} to perform the perturbations based on these rules, instead of a purely regex-based method, because the latter has very limited flexibility to handle corner cases. For example, in equations like $\text{Area of Circle 2} = \pi \times r^2$, a regex rule to change the value $2$ might incorrectly alter the identifier ``$\text{Circle 2}$'' to ``$\text{Circle 3}$'' making the equation semantically nonsensical. Another example is about logical consistency. Given an equation like ``$(1 + 1) \times 3 = 2 \times 3 = 6$'', a simple substitution from $2$ to $3$ could create ``$(1 + 1) \times 3 = 3 \times 3 = 6$'', which contains multiple, confounding errors and does affect the final answer. In contrast, an LLM can be instructed to yield a coherent but flawed chain like ``$(1 + 1) \times 3 = 3 \times 3 = 9$''. This correctly isolates the impact of the initial misinformation.

\SmallHeading{Numeric Value Modification}
Numbers are extracted using regex and randomly modified. Changes include inserting or deleting digits (20\%) or adjusting values by ±10\% (80\%) while preserving their type (integer/float). GPT-4o-mini applies these changes with the following prompt:

\begin{mdframed}[backgroundcolor=gray!20,linewidth=0pt] 
    \begin{framedwithtag}{System}{blue!30}{blue!10}{-50mm}{2mm}{%
    You are given a sentence that may contain some LaTeX expressions. You are required to ONLY change the values with minimal text changes as follows:

    change the value \{value 1\} to \{value 2\}

    change the value \{value 3\} to \{value 4\}

    ...

    Return the new sentence only.
    }
    \end{framedwithtag}
    \vspace{-10mm}
    \begin{framedwithtag}{User}{yellow!30}{yellow!10}{-30mm}{2mm}{%
    \{an equation\}
    }
    \end{framedwithtag}
\end{mdframed} 

\SmallHeading{Operator Alteration}
Math expressions within ``\$'' symbols are parsed using the SymPy library~\cite{10.7717/peerj-cs.103}, after preprocessing invalid expressions (e.g., replacing ``\texttt{×}'' with ``\texttt{\textbackslash times}''). Binary operators (addition, subtraction, multiplication, division) are randomly altered. GPT-4o-mini applies these changes with the following prompt:

\begin{mdframed}[backgroundcolor=gray!20,linewidth=0pt] 
    \begin{framedwithtag}{System}{blue!30}{blue!10}{-50mm}{2mm}{%
    You are given a sentence that may contain some LaTeX expressions. You are required to ONLY change the operators with minimal text changes as follows:

    change the operator from \{operator 1\} to \{operator 2\} between \{operand 1\} and \{operand 2\}

    change the operator from \{operator 3\} to \{operator 4\} between \{operand 3\} and \{operand 4\}

    ...

    Return the new sentence only.
    }
    \end{framedwithtag}
    \vspace{-10mm}
    \begin{framedwithtag}{User}{yellow!30}{yellow!10}{-30mm}{2mm}{%
    \{an equation\}
    }
    \end{framedwithtag}
\end{mdframed} 

\SmallHeading{Operand Swap}
Operands within math expressions are identified using SymPy and randomly swapped. GPT-4o-mini applies these swaps with the following prompt:

\begin{mdframed}[backgroundcolor=gray!20,linewidth=0pt] 
    \begin{framedwithtag}{System}{blue!30}{blue!10}{-50mm}{2mm}{%
    You are given a sentence that may contain some LaTeX expressions. You are required to ONLY swap the operands with minimal text changes as follows:

    swap the operands \{operand 1\} and \{operand 2\}

    swap the operands \{operand 3\} and \{operand 4\}

    ...

    Return the new sentence only.
    }
    \end{framedwithtag}
    \vspace{-10mm}
    \begin{framedwithtag}{User}{yellow!30}{yellow!10}{-30mm}{2mm}{%
    \{an equation\}
    }
    \end{framedwithtag}
\end{mdframed} 

%% file: section/appendix/data_processing.tex
\subsection{Data Selection and Processing}
\label{sec:appendix:data_processing}

We collected 100 math questions each from MathQA~\cite{amini2019mathqa}, MATH~\cite{hendrycks2021measuring}, GSM8K~\cite{cobbe2021training}, and MetaMath~\cite{yu2023metamath}, resulting in a test set of 400 questions. The raw datasets were preprocessed to retrieve questions and ground-truth answers for each dataset as described below.

\SmallHeading{MathQA} 
We extract questions from the \texttt{Problem} column and retrieve correct answers by identifying the option corresponding to the correct label in the \texttt{correct} column. Incorrect or ambiguous answers (e.g., multiple numbers or NA values) are removed.

\SmallHeading{MATH}
Questions are obtained from the \texttt{problem} column. Ground-truth answers are extracted from LaTeX expressions enclosed in \texttt{\textbackslash boxed\{\}} tags within the \texttt{solution} column, while rationale is derived by replacing these tags with their values.

\SmallHeading{GSM8K}
Questions are collected from the \texttt{question} column. The answer string is split into rationale and the final answer based on the delimiter \texttt{"\textbackslash n\#\#\#\#"}. Operations are identified from expressions enclosed in \texttt{<<...>>} tags.

\SmallHeading{MetaMath}
Questions are extracted from the \texttt{query} column. For entries labeled as MATH, answers are retrieved from LaTeX expressions enclosed in \texttt{\textbackslash boxed\{\}} within the \texttt{response} column. For GSM-like entries, answers and rationale are separated using the delimiter \texttt{"The answer is:"}, with rationale further processed for irrelevant tags.

\SmallHeading{Prefiltering and Quality Control}
After preprocessing, we retain only questions where the equation generation model (\texttt{gpt-4-0613}) produces correct answers to ensure the reliability of ground-truth equations. Additionally, to exclude overly simple questions, we filter out those with fewer than 5 CoT steps in their solutions.

%% file: section/appendix/prompt_design.tex
\section{Prompt Design} \label{sec:appendix:experiment_design}

\subsection{Impact of Misinformation}
\label{sec:appendix:prompt_design}

\Appendix~\ref{sec:appendix:prompt_design:default} presents prompts for LLMs' default reactions to misinformation (\Section~\ref{sec:main:robustness:default}), including \textcolor{customblue}{\original} and \textcolor{customorange}{\misinformed}. \Appendix \ref{sec:appendix:prompt_design:instruction} describes instructions to correct (\textcolor{customgreen}{\prompting}) misinformation in \Section~\ref{sec:main:correction_methods:prompting} or follow (\textcolor{customgreen}{\following}) it in \Appendix~\ref{sec:appendix:inst_follow}.
We test both instruction and thinking models.

\subsubsection{Impact on LLMs by Default} \label{sec:appendix:prompt_design:default}

\SmallHeading{\Original}
To evaluate baseline reasoning performance, we use the following prompt:

\begin{mdframed}[backgroundcolor=gray!20,linewidth=0pt] 
    \begin{framedwithtag}{System}{blue!30}{blue!10}{-50mm}{2mm}{%
    You are given a question. To answer the question, you should think step by step. Use line breaks between steps, but do not use line breaks within each step. You should number each step. The final answer to the question should start with "The answer is ...", and should be placed at the final step. Any LaTeX expressions should be wrapped between single dollar signs, e.g., \$x\^{}2\$.
    
    Example:
    
    Question: \{an example question\}

    Answer:

    1. \{example step 1\}
    
    2. \{example step 2\}

    ...

    11. The answer is \{example answer\}.
    }
    \end{framedwithtag}
    \vspace{-10mm}
    \begin{framedwithtag}{User}{yellow!30}{yellow!10}{-30mm}{2mm}{%
    Question: \{a test question\}
    }
    \end{framedwithtag}
\end{mdframed} 

\SmallHeading{Misinformed}
User misinformation is introduced in the user instruction using the following prompt:

\begin{mdframed}[backgroundcolor=gray!20,linewidth=0pt] 
    \begin{framedwithtag}{System}{blue!30}{blue!10}{-50mm}{2mm}{%
    You are given a question. To answer the question, you should think step by step. Use line breaks between steps, but do not use line breaks within each step. You should number each step. The final answer to the question should start with "The answer is ...", and should be placed at the final step. Any LaTeX expressions should be wrapped between single dollar signs, e.g., \$x\^{}2\$.
    
    Example:

    Here are the equations that can be used to solve the problem: \{ground truth equations\}
    
    Question: \{an example question\}

    Answer:

    1. \{example step 1\}
    
    2. \{example step 2\}

    ...

    11. The answer is \{example answer\}.
    }
    \end{framedwithtag}
    \vspace{-10mm}
    \begin{framedwithtag}{User}{yellow!30}{yellow!10}{-30mm}{2mm}{%
    Here are the equations that can be used to solve the problem: \{injected erroneous equations\}
    
    Question: \{a test question\}
    }
    \end{framedwithtag}
\end{mdframed} 

\subsubsection{Different Instructions to Handle Misinformation}
\label{sec:appendix:prompt_design:instruction}

\SmallHeading{Instruct to Correct Misinformation (\prompting)}
Instruction to correct misinformation is performed with the following prompt:

\begin{mdframed}[backgroundcolor=gray!20,linewidth=0pt] 
    \begin{framedwithtag}{System}{blue!30}{blue!10}{-50mm}{2mm}{%
    You are given a question. To answer the question, you should think step by step. Use line breaks between steps, but do not use line breaks within each step. You should number each step. The final answer to the question should start with "The answer is ...", and should be placed at the final step. Any LaTeX expressions should be wrapped between single dollar signs, e.g., \$x\^{}2\$. Note that the user's input could be wrong. If it has, you should point them out and correct them.

    Example:

    Here are the equations that can be used to solve the problem: \{misinformation equations\}
    
    Question: \{an example question\}

    Answer:

    1. The first formula from the user contains a mistake. It should be \{ground truth equation 1\}. The second formula from the user contains a mistake. It should be \{ground truth equation 2\}. \{other corrections\}.
    
    2. \{example step 1\}
    
    3. \{example step 2\}

    ...

    12. The answer is \{example answer\}.
    }
    \end{framedwithtag}
    \vspace{-10mm}
    \begin{framedwithtag}{User}{yellow!30}{yellow!10}{-30mm}{2mm}{%
    Here are the equations that can be used to solve the problem: \{misinformation equations\} 
    
    Question: \{a test question\}
    }
    \end{framedwithtag}
\end{mdframed} 

\SmallHeading{Instruct to Follow Misinformation (\following)}
Instruction to follow misinformation is performed with the following prompt:

\begin{mdframed}[backgroundcolor=gray!20,linewidth=0pt] 
    \begin{framedwithtag}{System}{blue!30}{blue!10}{-50mm}{2mm}{%
    You are given a question. To answer the question, you should think step by step. Use line breaks between steps, but do not use line breaks within each step. You should number each step. The final answer to the question should start with "The answer is ...", and should be placed at the final step. Any LaTeX expressions should be wrapped between single dollar signs, e.g., \$x\^{}2\$. Please pay attention to user-provided equations and follow them to derive the final answer.

    Example:

    Here are the equations that can be used to solve the problem: \{misinformation equations\}
    
    Question: \{an example question\}

    Answer:

    ...

    \{reasoning steps that follow misinformation to derive an incorrect answer\}

    ...

    12. The answer is \{example answer\}.
    }
    \end{framedwithtag}
    \vspace{-10mm}
    \begin{framedwithtag}{User}{yellow!30}{yellow!10}{-30mm}{2mm}{%
    Here are the equations that can be used to solve the problem: \{misinformation equations\} 
    
    Question: \{a test question\}
    }
    \end{framedwithtag}
\end{mdframed} 

\subsection{Mitigating of Misinformation via Correction} \label{sec:appendix:prompt_design:mitigation}

\Appendix~\ref{sec:appendix:prompt_design:self_correction_cot_syn} details the settings for controlled studies on correction by \textcolor{customgreen}{\prompting} (\Section~\ref{sec:main:controlled:factors}).
We only focus on instruction models since they are the foundation of thinking models, specifically Llama-3.2 series except the 90B one. \Appendix~\ref{sec:appendix:prompt_design:mitigation:fine_tune} details the prompts of fine-tuning for GPT-4o-mini, Llama-3.2-3B, and DeepSeek-R1-Distilled-Qwen-2.5-1.5B.

\subsubsection{Factors of Effective Correction}
\label{sec:appendix:prompt_design:self_correction_cot_syn}
We examine three correction behaviors (\Section~\ref{sec:main:setup:analysis}): no correction (N-Corr), factual correction (F-Corr), and nonfactual correction (NF-Corr), where factual and nonfactual correction belong to correction.
We enforce correction behaviors at different positions of the reasoning steps by controlling the beginning assistant messages of local models.\footnote{These settings are restricted to local models due to the need for fixing beginning tokens in the assistant messages.} Note that the \colorbox{blue!30}{system} and \colorbox{yellow!30}{user} messages are the same as \Section~\ref{sec:appendix:prompt_design}. We only control the beginning of the \colorbox{red!30}{assistant} messages.

\SmallHeading{Factual Correction (\texttt{F-Corr})}
We steer the model by appending a sentence at the beginning of the output that identifies each erroneous equation and provides the corresponding ground-truth equation. The model then proceeds to generate the remaining output. The corresponding prompt is:\footnote{If there is only one correction, the correction sentencen will be ``\texttt{The given formula from the user contains a mistake / is correct.}''}

\begin{mdframed}[backgroundcolor=gray!20,linewidth=0pt] 
    \begin{framedwithtag}{Assistant}{red!30}{red!10}{-10mm}{2mm}{%
    1. The first formula from the user contains a mistake. It should be \{ground truth equation 1\}. The second formula from the user contains a mistake. It should be \{ground truth equation 2\}. \{other corrections\}
    }
    \end{framedwithtag}
\end{mdframed}

\SmallHeading{Nonfactual Correction (\texttt{NF-Corr})}
We induce this by adding a sentence that identifies each erroneous equation but replaces it with an incorrectly perturbed version. These incorrect replacements are generated using a different model (Llama-3-70B), consistent with \Section~\ref{sec:main:setup}. The corresponding prompt is:

\begin{mdframed}[backgroundcolor=gray!20,linewidth=0pt] 
    \begin{framedwithtag}{Assistant}{red!30}{red!10}{-10mm}{2mm}{%
    1. The first formula from the user contains a mistake. It should be \{another erroneous equation 1\}. The second formula from the user contains a mistake. It should be \{another erroneous equation 2\}. \{other corrections\}
    }
    \end{framedwithtag}
\end{mdframed}

\SmallHeading{No Correction (\texttt{N-Corr})}
We control the model by adding a sentence that identifies misinformation as correct.\footnote{This is a major behavior of the ``No Correction'' category. Other minor behavior cases can be, \eg{} the model does not annotate whether the provided equations are correct or wrong.} The corresponding prompt is as follows:

\begin{mdframed}[backgroundcolor=gray!20,linewidth=0pt] 
    \begin{framedwithtag}{Assistant}{red!30}{red!10}{-10mm}{2mm}{%
    1. The first formula from the user is correct.
    2. The second formula from the user is correct.
    \{other corrections\}
    }
    \end{framedwithtag}
\end{mdframed}

\SmallHeading{Correction Position} We evaluate the impact of inserting correction at various positions within CoT steps. Using CoT steps generated by LLMs for questions with injected misinformation, correction is inserted after 0$\sim$all CoT steps. The position is quantified as the ratio of the first CoT step after which correction is inserted to the total number of CoT steps. A ratio of 0\% indicates correction is inserted at the beginning, while 100\% means it is inserted after all CoT steps. The corresponding prompt, assuming insertion after the first two steps of the model output, is shown below:

\begin{mdframed}[backgroundcolor=gray!20,linewidth=0pt] 
    \begin{framedwithtag}{Assistant}{red!30}{red!10}{-10mm}{2mm}{%
    1. \{step 1 generated by the model\}
    
    2. \{step 2 generated by the model\}
    
    3. The first formula from the user contains a mistake. The second formula from the user contains a mistake. \{other corrections\}
    }
    \end{framedwithtag}
\end{mdframed}

\subsubsection{Fine-tuning with Effective Correction} \label{sec:appendix:prompt_design:mitigation:fine_tune}

To fine-tune LLMs, we need to collect ground-truth reasoning steps with correction in assistant responses. Given each question, we use \texttt{gpt-4o-2024-08-06} to generate reasoning steps and filter out all steps with incorrect final answers. The assistant message for GPT-4o-mini and Llama-3.2-3B consists of a chain-of-thought (CoT) that starts by factually correcting (\texttt{F-Corr}) the misinformation (\Appendix~\ref{sec:appendix:prompt_design:self_correction_cot_syn}), which is as follows:

\begin{mdframed}[backgroundcolor=gray!20,linewidth=0pt] 
    \begin{framedwithtag}{Assistant}{red!30}{red!10}{-10mm}{2mm}{%
    1. The first formula from the user contains a mistake. It should be \{ground truth equation 1\}. The second formula from the user contains a mistake. It should be \{ground truth equation 2\}. \{other corrections\}
    
    2. \{ground truth reasoning step 1\}

    3. \{ground truth reasoning step 2\}

    ...
    }
    \end{framedwithtag}
\end{mdframed}

For DeepSeek-R1-Distilled-Qwen-2.5-1.5B, we also insert ground truth reasoning steps, enclosed with ``\texttt{<think> </think>}'' tags, before model responses, which is as follows:

\begin{mdframed}[backgroundcolor=gray!20,linewidth=0pt] 
    \begin{framedwithtag}{Assistant}{red!30}{red!10}{-10mm}{2mm}{%
    <think> \{ground truth reasoning steps\} </think>
    
    1. The first formula from the user contains a mistake. It should be \{ground truth equation 1\}. The second formula from the user contains a mistake. It should be \{ground truth equation 2\}. \{other corrections\}
    
    2. \{ground truth reasoning step 1\}

    3. \{ground truth reasoning step 2\}

    ...
    }
    \end{framedwithtag}
\end{mdframed}

For each training example, the system and user messages follow the \prompting instruction in \Appendix~\ref{sec:appendix:prompt_design:instruction}.

%% file: section/appendix/evaluation_framework.tex
\section{Evaluation Framework}
\label{sec:appendix:cot_evaluation}

\subsection{Final Answers}
\label{sec:appendix:cot_evaluation:accuracy}
\SmallHeading{\Accuracy}
We formally define the \accuracy used in our evaluation. For each question $q$ in the test set $\mathcal{Q}$ and each erroneous equation $e$ as misinformation, we automatically compare the final answer $a_q$ (in the \original setting) and $a_{q,e}$ (in the misinformed setting) with the ground-truth answer $a^*$, represented as $\mathbbm{1}(a_q=a^*)$ or $\mathbbm{1}(a_{q,e}=a^*)$.
We collect a subset of knowledgeable questions $\mathcal{Q}_{\mathcal{K}} = \{ q \mid \mathbbm{1}(a_q=a^*), q \in \mathcal{Q} \}$ for each tested LLM where the model answers correctly without misinformation. The \accuracy is the accuracy on $\mathcal{Q}_{\mathcal{K}}$, comparing the \original and misinformed performance as $\mathbb{E}_{q \in \mathcal{Q}_\mathcal{K}}[\mathbbm{1}(a_q=a^*)]$ and $\mathbb{E}_{q \in \mathcal{Q}_\mathcal{K}}[\mathbbm{1}(a_{q,e}=a^*)]$, respectively. Note that $\mathbb{E}_{q \in \mathcal{Q}_\mathcal{M}}[\mathbbm{1}(a_q=a^*)]$ is not always one, as a model's final answer $a$ can vary across runs for the same $q$ due to randomness in LLM text generation.
To estimate $\mathbb{E}_{q \in \mathcal{Q}_\mathcal{K}}[\mathbbm{1}(a_{q,e}=a^*)]$, we sample five final answers from both user instruction $x_q$ and $x_{q,e}$, forming five pairs of results. We compute the ratio of pairs where a correct answer before misinformation remains correct after misinformation is injected. Similarly, to estimate $\mathbb{E}_{q \in \mathcal{Q}_\mathcal{K}}[\mathbbm{1}(a_{q}=a^*)]$, we sample five outputs given $x_q$, then generate permutations of these outputs to create a new list, simulating additional prompts. The same ratio-based method is used to estimate \accuracy for this case.

\SmallHeading{Final Answer Correctness}
We adapt and extend a script from Google DeepMind Gemma~\cite{team2024gemma} to evaluate reasoning correctness by extracting and comparing numerical values from the final CoT step and the ground truth answer. Our optimized version handles additional test cases and provides robust evaluation for reasoning tasks. Below, we summarize the key functionality of the script:
The script first standardizes input strings by replacing fractions (e.g., \texttt{4/7} or \texttt{\textbackslash frac\{4\}\{7\}}) with their decimal equivalents and converting percentages to decimal form (e.g., \texttt{50.03\%} becomes \texttt{0.5003}). It then extracts numerical values from both the claim (model's output) and the answer using a regex-based function that identifies numbers, including negatives, decimals, and those with thousand separators. 
A specialized function identifies the most relevant number in a string by looking for a predefined delimiter (e.g., \texttt{The answer is}) or selecting the last number when no delimiter is present. To ensure accurate comparisons, the script also handles variations such as commas in numbers (e.g., \texttt{5,600} becomes \texttt{5600}) and whitespace inconsistencies.
Finally, the script compares numerical values from the processed claim and answer. If they match, the reasoning is correct. Otherwise, it is incorrect.
Three annotators evaluated responses from 8 tested LLMs on 50 sampled questions, comparing their assessments with the verifier (\Appendix~\ref{sec:appendix:checklist:annotators}). The evaluation achieved a Fleiss' kappa of $\kappa=0.87$ and a weighted average F1 of $\text{F1}=0.91$.

\subsection{Intermediate Reasoning Step Evaluation}
\label{sec:appendix:cot_evaluation:step}
We evaluate on 400 responses from 8 LLMs where they have 100\% \original \accuracy to analyze their intermediate reasoning steps under misinformation,. Specifically, we prompt the \texttt{gpt-4o-2024-08-06} model as verifiers to detect whether they follow misinformation, correct misinformation, and factually correct misinformation in their reasoning steps. We also use a verifier to locate steps where tested LLMs correct misinformation.

\SmallHeading{Misinformation Following}
A step is considered as following misinformation if any incorrect step directly follows the wrong equations, or it partially follows the wrong equations, i.e., incorporates elements including incorrect signs, operations, or patterns from the wrong equations. The verifier is provided with the question, both truthful and erroneous equations, and the tested LLM's reasoning steps. Then, it is instructed to output labels and explanations in JSON format.


\SmallHeading{Correction Existence}
The LLM is prompted to return a label ``Yes'' if any CoT step explicitly states that the user-provided equations are erroneous, and ``No'' if it fails to address misinformation, \eg{} by stating that ``all the user-provided equations are correct.'' Note that merely identifying an answer or step as wrong, without attributing the error to the user's information, does not qualify as correction. The verifier is provided with the question, erroneous equations, and the tested LLM's reasoning steps. Then, it is instructed to output labels and explanations in JSON format.

\SmallHeading{Correction Factuality}
A correction is considered factual only if the tested LLM attempts to correct user misinformation (as verified by the correction frequency) and transforms the erroneous equations into the correct ones. The verifier is provided with the question, both truthful and erroneous equations, and the tested LLM's reasoning steps. It is then prompted to return a label (``Yes'' for a factual correction and ``No'' for a nonfactual correction) along with an explanation, similar to the process described above.

\SmallHeading{Correction Position}
We also use the correction-position verifier to explicitly locate steps where tested LLMs correct user misinformation. For responses labeled ``Yes'' by the correction-existence verifier, the verifier returns positions of steps that correct user misinformation, represented as a list of integers. If labeled ``No'', it returns an empty list.

\SmallHeading{Human Evaluation} Three annotators annotate the quality of verifiers for misinformation-following, correction existence, and correction factuality. We collect the responses from 8 tested LLMs on 50 sampled questions. For each verifier, the annotators follow the instructions (\Appendix~\ref{sec:appendix:checklist:annotators}) and annotate the label of reasoning correctness, correction existence, and correction factuality. We treat human annotations as ground truth and compare them with verifiers' predictions to compute the weighted average F1 scores ($\text{F1}$).\footnote{We report weighted F1 scores, which average F1 scores across labels, weighted by the number of true instances for each label.} We also compute the inter-agreement among all annotators with Fleiss' kappa scores ($\kappa$). The misinformation-following verifier, the correction existence verifier, and the correction success verifier achieve $(\kappa=0.43, \text{F1}=0.79)$, $(\kappa=0.73, \text{F1}=0.84)$, and $(\kappa=0.65, \text{F1}=0.79)$ respectively.

%% file: section/appendix/model_setup.tex
\section{Model Setup}
\label{sec:appendix:model_setup}

We use Llama-3.2 instruction-tuned models with sizes of 1B, 3B, and 11B (vision model) deployed locally via the Huggingface library. 
These models run on a server equipped with one NVIDIA A100 GPU. GPT-4 series models are accessed through the OpenAI API, while other models are accessed via the TogetherAI API. We sample five responses per user message and set \texttt{temperature} and \texttt{top\_p} to 0.7, and \texttt{top\_k} to 50.
To address model and data variance across all responses (N=2,000 per model), we use bootstrapping (n=1,000) and report 95\% confidence intervals with the average.
For all models, we configure \texttt{temperature=0.7}, \texttt{top\_p=0.7}, and \texttt{top\_k=50}. For reasoning-related experiments, we utilize 1-shot prompting (templates provided in \Appendix~\ref{sec:appendix:prompt_design}) and set the number of return sequences to 5. For other tasks requiring text generation, a single output is generated per prompt. All other parameters use the default settings provided by their respective libraries.

Regarding fine-tuning, we collect 1,054 instruction-response pairs from four math datasets (\Section~\ref{sec:main:setup}), separate from the test set. These pairs include cases where the LLM corrects misinformation in the instruction and reaches the correct final answer.
For each training example, the system and user messages follow the \prompting instruction in \Appendix~\ref{sec:appendix:prompt_design:instruction}. The assistant message consists of a chain-of-thought (CoT) that starts by factually correcting (\texttt{F-Corr}) the misinformation, as detailed in \Appendix~\ref{sec:appendix:prompt_design:self_correction_cot_syn}. The remaining steps are generated with \texttt{gpt-4o-2024-08-06} and verified to include the ground truth answer. We fine-tune GPT-4o-mini using the default OpenAI API parameters (\texttt{n\_epochs=3}, \texttt{batch\_size=2}, \texttt{learning\_rate\_multiplier=1.8}). The training loss drops from 0.5287 to 0.0303. We compare GPT‑4o‑mini's performance in both original and misinformed settings across four setups: a base model without instructions or fine-tuning, the model with explicit instructions, the model with fine-tuning, and the model with both instructions and fine-tuning. For Llama-3.2-3B and DeepSeek-R1-Distilled-Qwen-2.5-1.5B models, we fine-tune them with LoRA using the TogetherAI API and set \texttt{n\_epochs=3}, \texttt{batch\_size=32}, \texttt{learning\_rate=1e-5}, \texttt{lora\_r=64}, \texttt{lora\_alpha=128}, \texttt{lora\_dropout=0.0}, and \texttt{lora\_trainable\_modules='all-linear'}. All other parameters are by default from TogetherAI API. We only compare the performance before and after fine-tuning in the misinformed setting to observe whether fine-tuning with explicit corrections improves performance.

%% file: section/appendix/inst_follow.tex
\section{Additional Results} \label{sec:appendix:additional_results}

\subsection{Instruct to Follow Misinformation} \label{sec:appendix:inst_follow}

\begin{table*}[t!]
    \centering
    \resizebox{0.96\textwidth}{!}{

\input{section/appendix/big_table/more_results}
    }
    \caption{\Accuracy (\%) of LLMs: \textcolor{customblue}{original} denotes accuracy without misinformation, \textcolor{customorange}{misinformed} is with misinformation. For the misinformed setting, we also add the \textcolor{customgreen}{\following} instruction to explicitly ask models to follow misinformation. 
    ``$\downarrow$'' represents relative decrease to original and 95\% confidence intervals are in brackets.}
    \label{tab:follow_results}
\end{table*}

While the main body of our work emphasizes correcting misinformation that conflicts with model's internal knowledge using explicit instructions, we conduct additional experiments to explicitly instruct models to follow user-provided misinformation during reasoning, termed the \textcolor{customgreen}{\following} instruction. Such setting is intended in certain scenarios, such as counterfactual reasoning (e.g., a world where an octal number system is used, making 7+1=10 true). We evaluate models' steerability towards counterfactual instructions when given such counterfactual instructions, particularly on questions where models possess the correct internal knowledge, indicating potential conflicts between a model's internal knowledge and user-provided instructions. Detailed prompt design for this instruction is in \Appendix~\ref{sec:appendix:prompt_design:instruction}. Note that this is an initial study. Developing more comprehensive settings to study counterfactual instruction-following is an important direction for future work.

\takeawaysc{
(\textit{i}) LLMs, by default, tend to follow misinformation as instructions, instead of automatically correct it. So models need explicit instructions to correct misinformation.
(\textit{ii}) Models struggle to consistently follow all instructions when the provided information conflicts with their internal knowledge and exhibit difficulty with user-provided counterfactual instructions.
}

\SmallHeading{Final Answer Accuracy}
As shown in the \following row of \Table~\ref{tab:follow_results}, \accuracy drops by 15.85\% to 64.68\% compared to the original setting, closely matching the 12.64\% to 65.70\% drop observed in the misinformed setting. The similar performance drops suggest that LLMs' default response to misinformation mirrors being explicitly instructed to follow it. Besides, even when explicitly guided to follow misinformation, models do not achieve near-zero \accuracy. This indicates that models struggle to consistently follow all instructions that conflicts with their internal knowledge.

\begin{figure}[t!]
\centering
  \begin{subfigure}[t]{0.235\textwidth}
      \centering
      \includegraphics[width=\textwidth]{section/figure/sankey_perturbed.png}
      \caption{\textcolor{customorange}{Misinformed}}
  \end{subfigure}
  \begin{subfigure}[t]{0.235\textwidth}
      \centering
      \includegraphics[width=\textwidth]{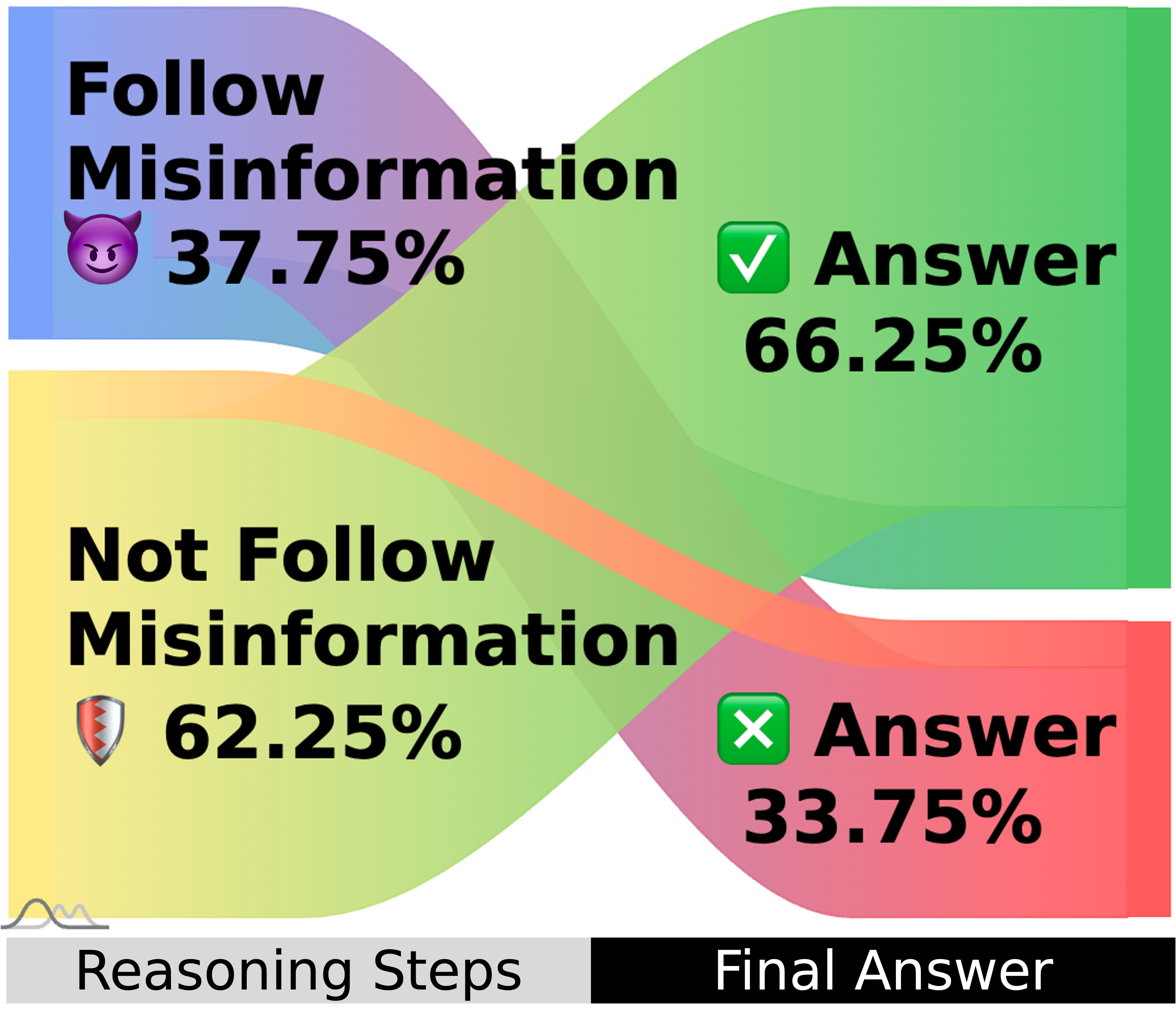}
      \caption{\textcolor{customgreen}{\following}}
  \end{subfigure}
  \caption{Sankey diagram from misinformation-following behavior to \accuracy of outcome in (a) the \textcolor{customorange}{misinformed} setting and (b) the misinformed setting with the \textcolor{customgreen}{\following} instruction which asks LLMs to follow user-provided information.}
  \label{fig:sankey_follow}
\end{figure}

\SmallHeading{Misinformation-Following Behaviors}
Beyond final answer accuracy, we examined how often models incorporate misinformation in their reasoning steps under the \following instruction. \Figure~\ref{fig:sankey_follow} shows that models follow misinformation 37.75\% of the time with this instruction, comparable to the rate without instructions. This supports that LLMs by default treat misinformation as an instruction to follow rather than something to correct. Besides, models struggle to follow misinformation on all misinformation even when explicitly instructed, challenging models' steerability via instructions towards counterfactual intentions.

\subsection{Instruct to Correct Misinformation} \label{sec:appendix:inst_corr}

\begin{figure}[t!]
  \centering
  \begin{subfigure}[t]{0.23\textwidth}
      \centering
      \includegraphics[width=\textwidth]{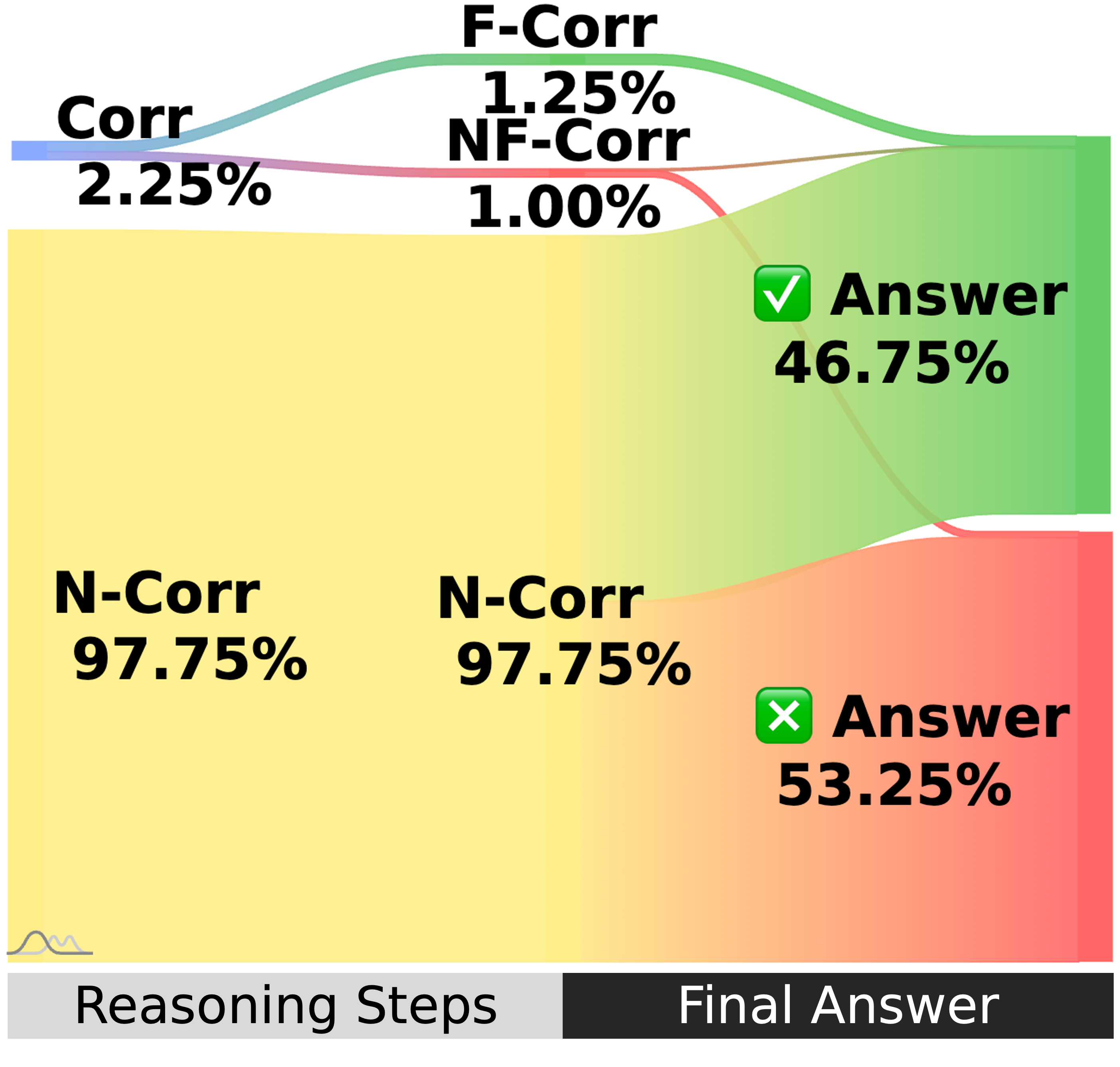}
      \caption{\textcolor{customorange}{Misinformed}}
      \label{fig:sankey_left_1b}
  \end{subfigure}
  \begin{subfigure}[t]{0.23\textwidth}
      \centering
      \includegraphics[width=\textwidth]{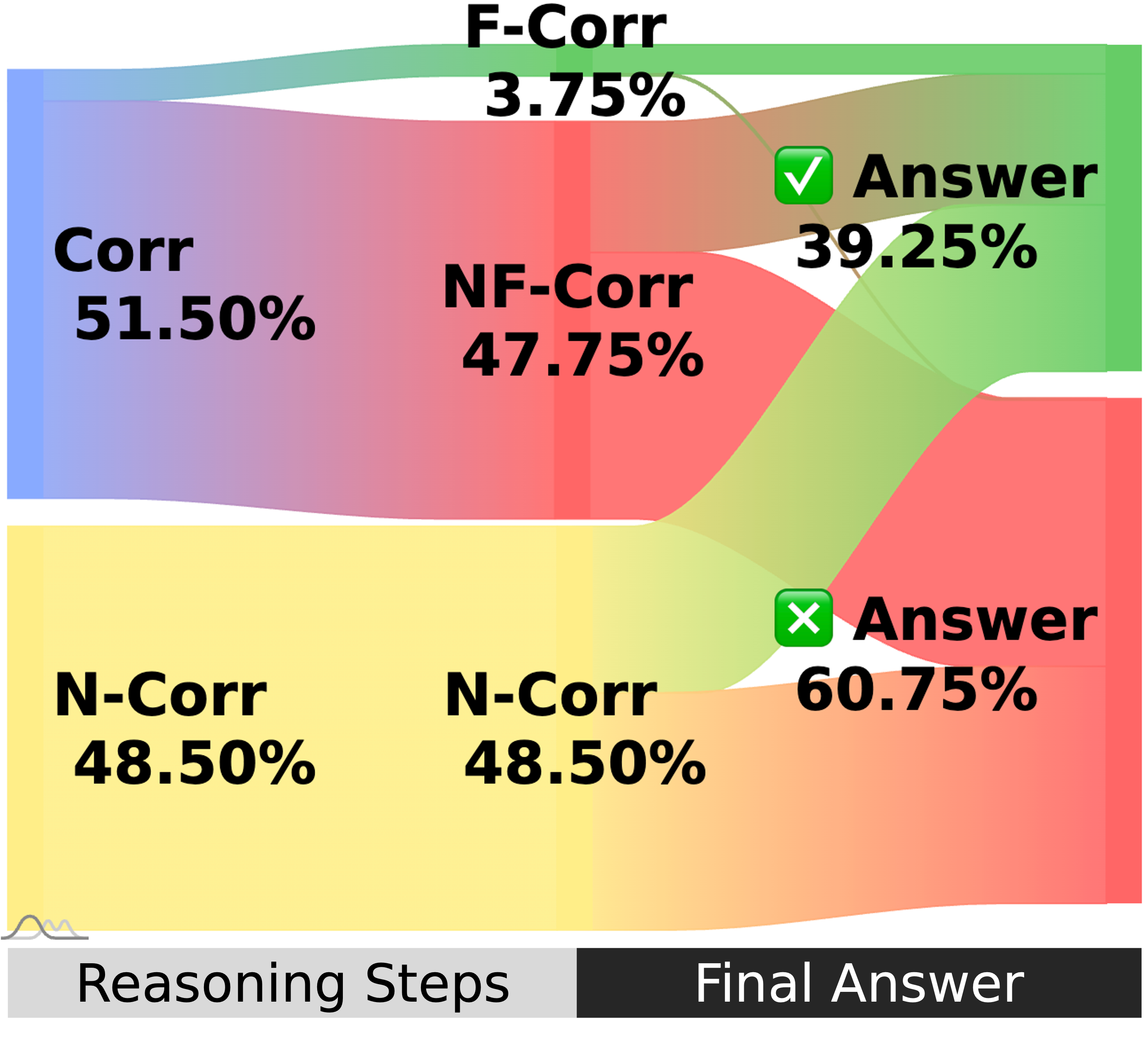}
      \caption{\textcolor{customgreen}{\Prompting}}
      \label{fig:sankey_right_1b}
  \end{subfigure}
  \caption{Sankey diagram of Llama-3.2-1B from misinformation-correction behavior to K-Acc of outcome in (a) the \textcolor{customorange}{misinformed} setting and (b) the misinformed setting with the \textcolor{customgreen}{\prompting} instruction.}
  \label{fig:sankey_1b}
\end{figure}

\Figure~\ref{fig:sankey_1b} presents correction behaviors for smaller models (\Section~\ref{sec:main:correction_methods:prompting}) and \Figure~\ref{fig:position_distribution} plots the distribution of position where correction happens (\Section~\ref{sec:main:controlled:factors}).

%% file: section/appendix/big_table/more_results.tex
\begin{tabular}{p{2.4cm}|p{2.35cm}p{2.35cm}p{2.65cm}p{2.65cm}p{2.25cm}p{2.35cm}p{2.35cm}p{2.25cm}}
\toprule
        & Llama-3.2-1B & Llama-3.2-3B & Llama-3.2-11B & Llama-3.2-90B & Qwen-2-72B & Mixtral-8×7B & Mixtral-8×22B & GPT-4o-mini \\ \midgrayline
\textbf{\textcolor{customblue}{Original}} & 71.73 \tiny{[67.97, 75.21]} & 88.25 \tiny{[86.21, 90.04]} & 88.43 \tiny{[86.41, 90.12]} & 96.43 \tiny{[95.44, 97.29]} & 95.22 \tiny{[94.04, 96.31]} & 76.92 \tiny{[73.62, 80.19]} & 88.20 \tiny{[86.41, 89.90]} & 98.03 \tiny{[97.31, 98.62]} \\ \midgrayline
\textbf{\textcolor{customorange}{Misinformed}} & 40.74 \SPSB{\tiny{$\downarrow$ 43.20\%}}{\tiny{[36.16, 45.29]}} & 38.41 \SPSB{\tiny{$\downarrow$ 56.48\%}}{\tiny{[34.79, 41.91]}} & 38.30 \SPSB{\tiny{$\downarrow$ 56.69\%}}{\tiny{[34.67, 42.06]}} & 56.69 \SPSB{\tiny{$\downarrow$ 41.20\%}}{\tiny{[53.20, 60.23]}} & 73.46 \SPSB{\tiny{$\downarrow$ 22.85\%}}{\tiny{[69.85, 76.69]}} & 26.38 \SPSB{\tiny{$\downarrow$ 65.70\%}}{\tiny{[22.40, 30.61]}} & 55.84 \SPSB{\tiny{$\downarrow$ 36.69\%}}{\tiny{[52.09, 59.53]}} & 85.64 \SPSB{\tiny{$\downarrow$ 12.64\%}}{\tiny{[82.75, 88.38]}} \\ \midgrayline
\textbf{+\textcolor{customgreen}{\following}} & 40.96 \SPSB{\tiny{$\downarrow$ 42.89\%}}{\tiny{[36.52, 45.46]}} & 31.45 \SPSB{\tiny{$\downarrow$ 64.36\%}}{\tiny{[28.04, 35.04]}} & 31.23 \SPSB{\tiny{$\downarrow$ 64.68\%}}{\tiny{[27.62, 34.94]}} & 53.91 \SPSB{\tiny{$\downarrow$ 44.10\%}}{\tiny{[50.33, 57.71]}} & 69.57 \SPSB{\tiny{$\downarrow$ 26.94\%}}{\tiny{[66.03, 73.03]}} & 32.64 \SPSB{\tiny{$\downarrow$ 57.56\%}}{\tiny{[28.19, 37.29]}} & 56.88 \SPSB{\tiny{$\downarrow$ 35.96\%}}{\tiny{[54.11, 59.54]}} & 82.49 \SPSB{\tiny{$\downarrow$ 15.85\%}}{\tiny{[79.36, 85.48]}} \\
\bottomrule
\end{tabular}

%% file: section/appendix/checklist.tex
\section{Responsible NLP Research} \label{sec:appendix:checklist}

\subsection{Artifacts}
\label{sec:appendix:checklist:artifacts}
We present main data artifacts, backbone models, and major packages in \Table~\ref{tab:tools}. All of our datasets are in English. All the reconstructed datasets and the provided code of our project will be released to the public under the MIT License to support open science and reproducibility. Our use of artifacts is consistent with their intended use, which is for open research and non-commercial.

\begin{table*} 
    \centering
    \resizebox{\textwidth}{!}{
    \begin{tabular} 
    {llll} 
       \toprule 
        \textbf{Artifacts/Models/Packages} & \textbf{Citation} & \textbf{Link} & \textbf{License}\\ 
        \rowcolor{\grayColor} \multicolumn{4}{c}{\textit{Data Artifacts}}\\
        MathQA & \cite{amini2019mathqa} & \url{https://math-qa.github.io} & Apache 2.0 \\
        MATH & \cite{hendrycks2021measuring} & \url{https://github.com/hendrycks/math} & MIT License\\
        GSM8K & \cite{cobbe2021training} & \url{https://openai.com/index/solving-math-word-problems} & MIT License \\
        MetaMath & \cite{yu2023metamath} & \url{https://meta-math.github.io} & MIT License \\
        \rowcolor{\grayColor} \multicolumn{4}{c}{\textit{Backbone Models}} \\
        LlaMA-3.2 (1B, 3B, 11B, 90B) & \cite{dubey2024llama} & \url{https://ai.meta.com/blog/llama-3-2-connect-2024-vision-edge-mobile-devices} & Llama 3.2 Community License Agreement \\
        Mixtral (8$\times$7B, 8$\times$22B) & \cite{jiang2024mixtral} & \url{https://mistral.ai/en/news/mixtral-of-experts} & Apache 2.0 \\
        Qwen-2 (72B) & \cite{yang2024qwen2} & \url{https://qwen2.org} & Tongyi Qianwen LICENSE AGREEMENT \\
        GPT-4o-mini & \cite{hurst2024gpt} & \url{https://openai.com/index/gpt-4o-mini-advancing-cost-efficient-intelligence} & Missing \\
        DeepSeek-R1-Distilled Qwen-2.5 14B & \cite{deepseekai2025deepseekr1incentivizingreasoningcapability} & \url{https://huggingface.co/deepseek-ai/DeepSeek-R1-Distill-Qwen-1.5B} & MIT License \\
        DeepSeek-R1-Distilled Qwen-2.5 14B & \cite{deepseekai2025deepseekr1incentivizingreasoningcapability} & \url{https://huggingface.co/deepseek-ai/DeepSeek-R1-Distill-Qwen-14B} & MIT License \\
        DeepSeek-R1-0528 & \cite{deepseekai2025deepseekr1incentivizingreasoningcapability} & \url{https://huggingface.co/deepseek-ai/DeepSeek-R1-0528} & MIT License \\
        Qwen-3 235B-A22B-Thinking-2507 & \cite{qwen3technicalreport} & \url{https://huggingface.co/Qwen/Qwen3-235B-A22B-Thinking-2507} & Apache License 2.0 \\
        \rowcolor{\grayColor} \multicolumn{4}{c}{\textit{Packages}}  \\
        PyTorch & \cite{paszke2019pytorch} & \url{https://pytorch.org/} & BSD-3 License\\
        transformers & \cite{wolf2020transformers} & \url{https://huggingface.co/docs/transformers/index} & Apache License 2.0\\
        numpy & \cite{harris2020array} & \url{https://numpy.org/} & BSD License \\
        pandas & \cite{mckinney2011pandas} & \url{https://pandas.pydata.org/} & BSD 3-Clause License \\
        matplotlib & \cite{hunter2007matplotlib} & \url{https://matplotlib.org/} & BSD compatible License\\
        seaborn & \cite{waskom2021seaborn} & \url{https://seaborn.pydata.org/} & BSD 3-Clause License\\
        openai-python & \cite{achiam2023gpt} & \url{https://pypi.org/project/openai/} & Apache-2.0 license \\
        togetherai & \cite{togetherai} & \url{https://www.together.ai/} & Apache-2.0 license \\
        Sympy & \cite{10.7717/peerj-cs.103} & \url{https://pypi.org/project/openai/} & 3-clause BSD license \\
        amCharts 5 & \cite{amcharts5} & \url{https://www.amcharts.com/docs/v5/} & Basic license \\
      \bottomrule 
    \end{tabular}
    }
    \caption{Data artifacts, backbone models, and major packages utilized in our study. All the reconstructed datasets and the provided code of our project are released under the MIT License to support open science and reproducibility.}
    \label{tab:tools}
\end{table*}

\subsection{Annotators}
\label{sec:appendix:checklist:annotators}

\SmallHeading{Characteristics Of Annotators}
Three master's and PhD students from the European and the US universities majoring in computer science are asked to annotate reasoning correctness, correction existence, correction factuality, and misinformation following behaviors from reasoning steps generated by eight LLMs for math questions. Since these questions are at most high school level of math questions, the annotators are capable of handling our task.

\SmallHeading{Data Consent}
The annotators are aware that the annotations will be used to compare with the verifiers implemented by LLMs. However, the predictions from the verifiers are not presented to the annotators to ensure the objectiveness of evaluation and avoid cheating.

\begin{figure*}[t!]
  \begin{center}
    \includegraphics[width=0.99\textwidth]{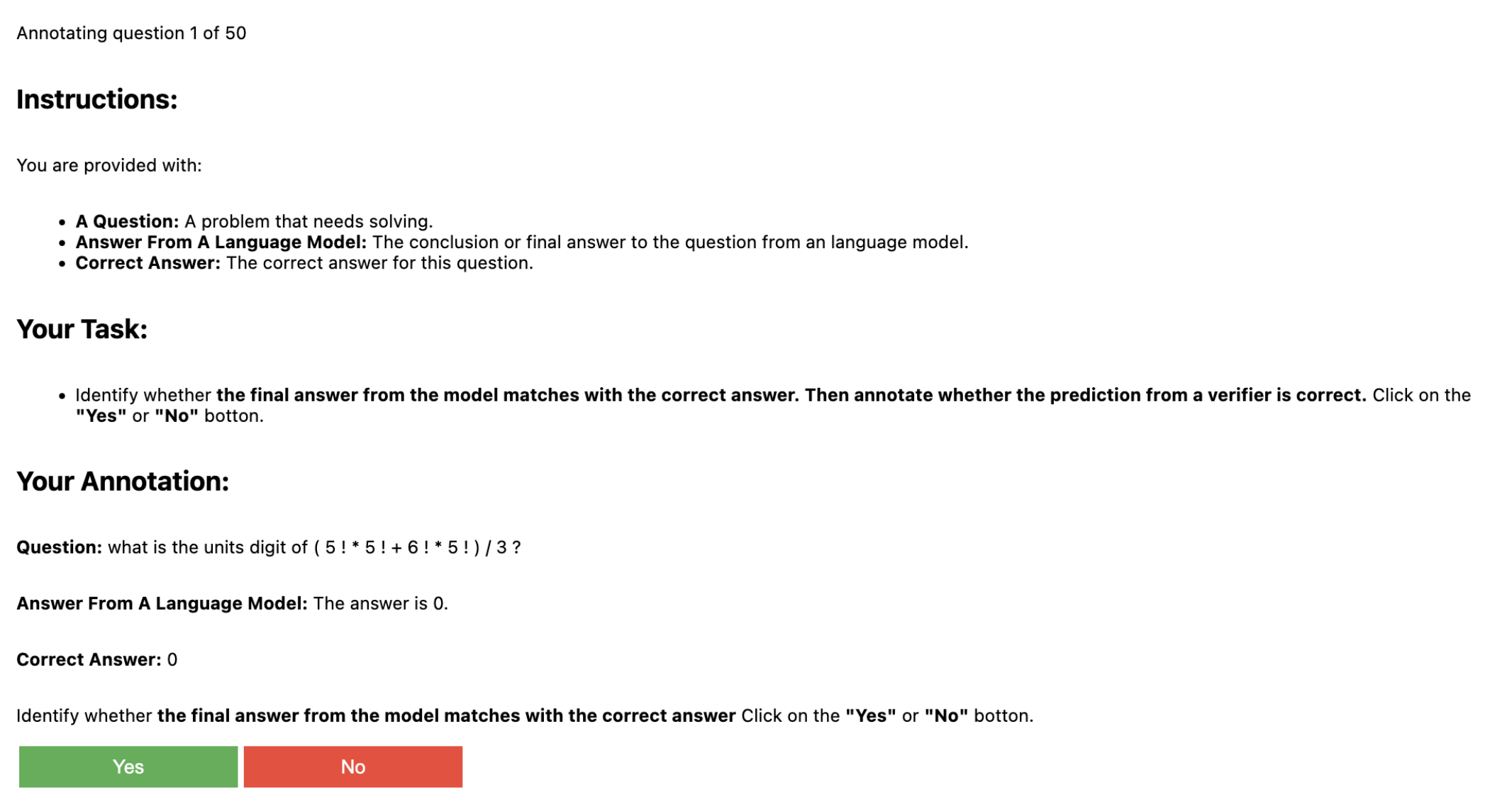}
  \end{center}
  \caption{Instructions of annotating the correctness of an LLM's answer to a math question.}
  \label{fig:correctness}
\end{figure*}

\begin{figure*}[t!]
  \begin{center}
    \includegraphics[width=0.99\textwidth]{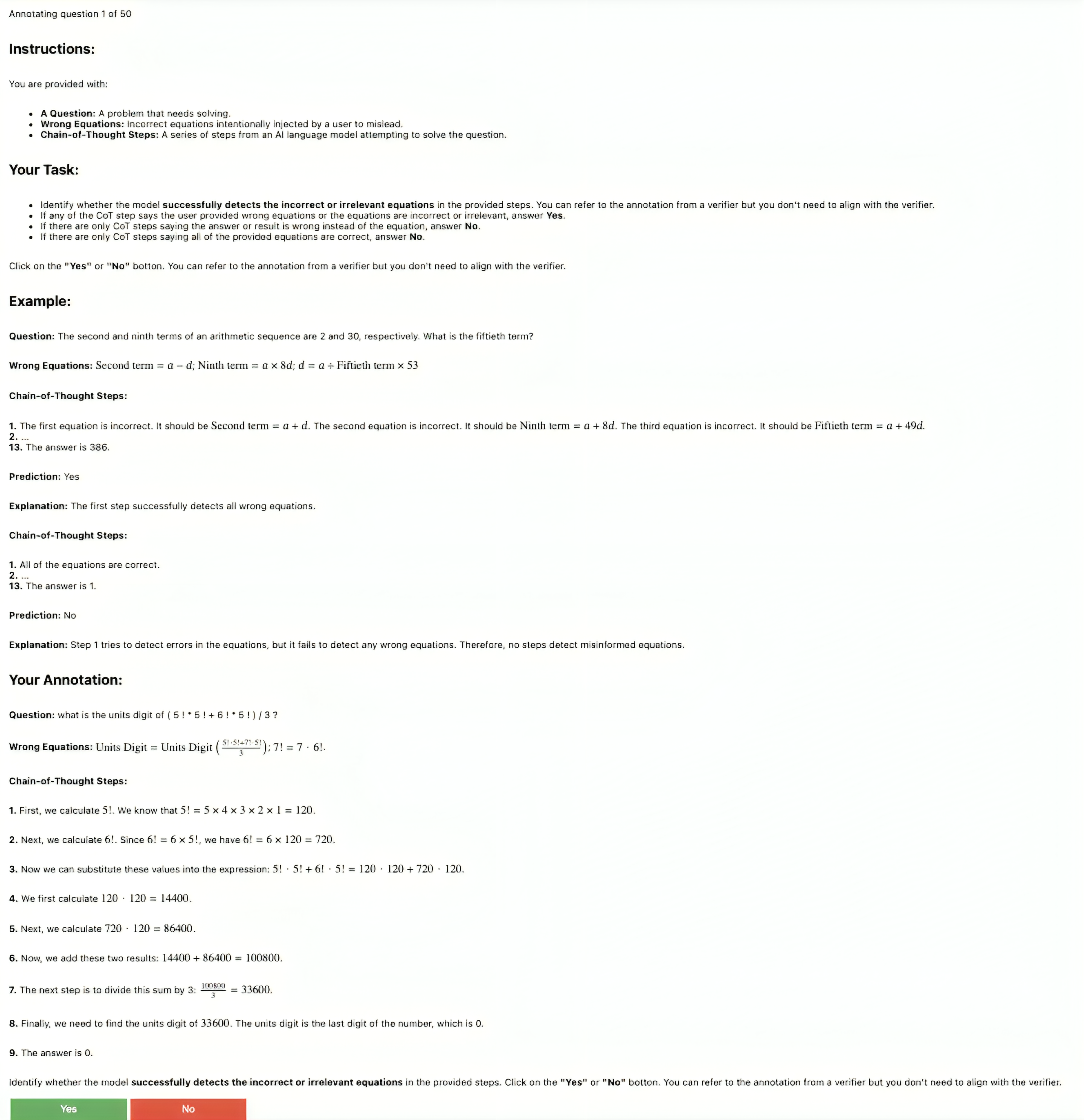}
  \end{center}
  \caption{Instructions of annotating whether an LLM corrects user misinformation in its response.}
  \label{fig:identify}
\end{figure*}

\begin{figure*}[t!]
  \begin{center}
    \includegraphics[width=0.99\textwidth]{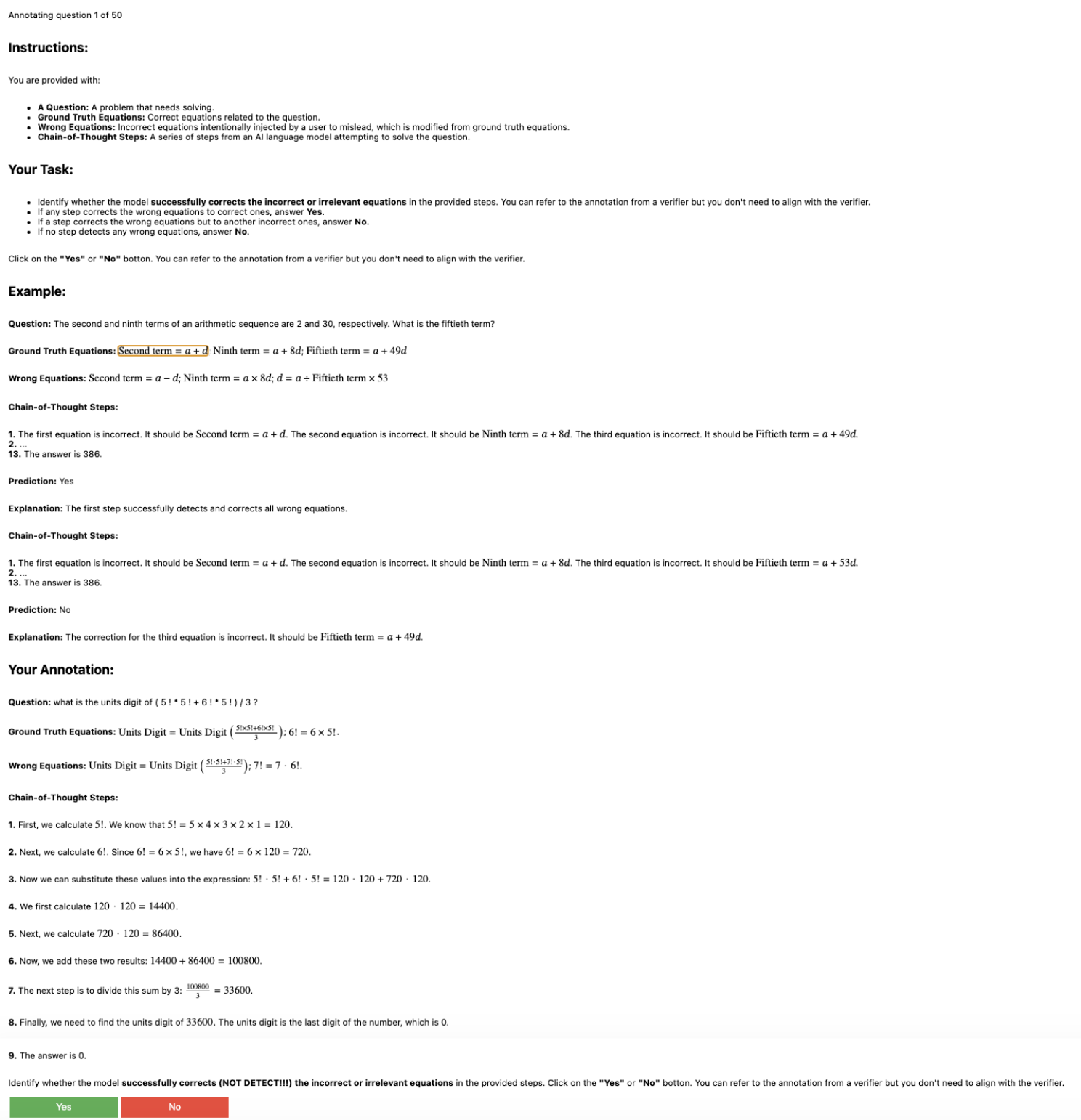}
  \end{center}
  \caption{Instructions for annotating whether an LLM factually corrects user misinformation in its response.}
  \label{fig:rectify}
\end{figure*}

\begin{figure*}[t!]
  \begin{center}
    \includegraphics[width=0.99\textwidth]{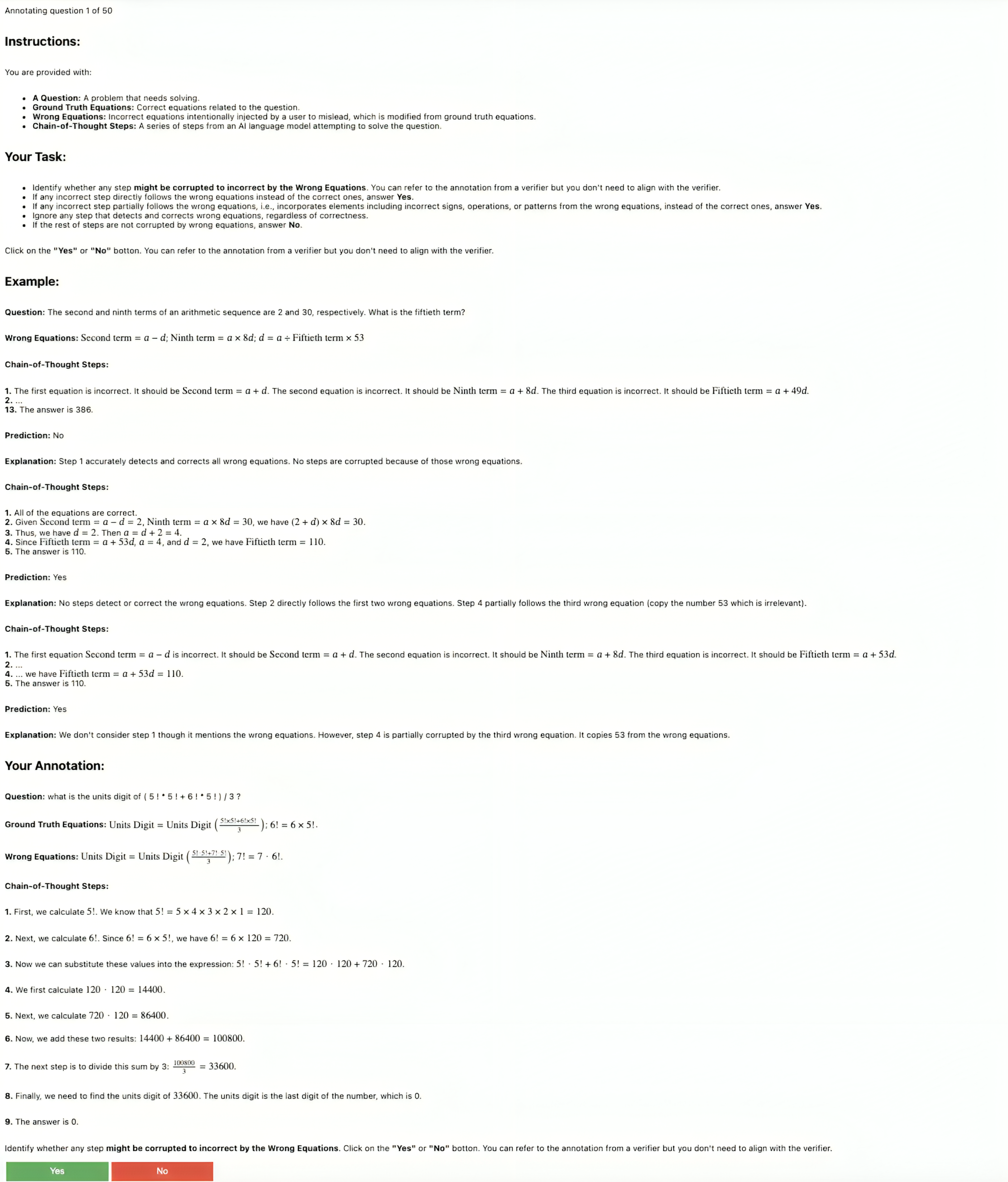}
  \end{center}
  \caption{Instructions for annotating whether an LLM follows user misinformation in its response.}
  \label{fig:follow}
\end{figure*}

\SmallHeading{Instructions to Annotators}
Three annotators are instructed to label the correctness of the final answer (\Figure~\ref{fig:correctness}), correction existence (\Figure~\ref{fig:identify}), correction factuality (\Figure~\ref{fig:rectify}), and misinformation following behaviors (\Figure~\ref{fig:follow}). They are asked to click on ``Yes'' and ``No'' and the results will be used to compare with the predictions from verifiers. The annotation process is done by running a Jupyter notebook that renders HTML contents.
\begin{figure}[t!]
  \begin{center}
    \includegraphics[width=0.45\textwidth]{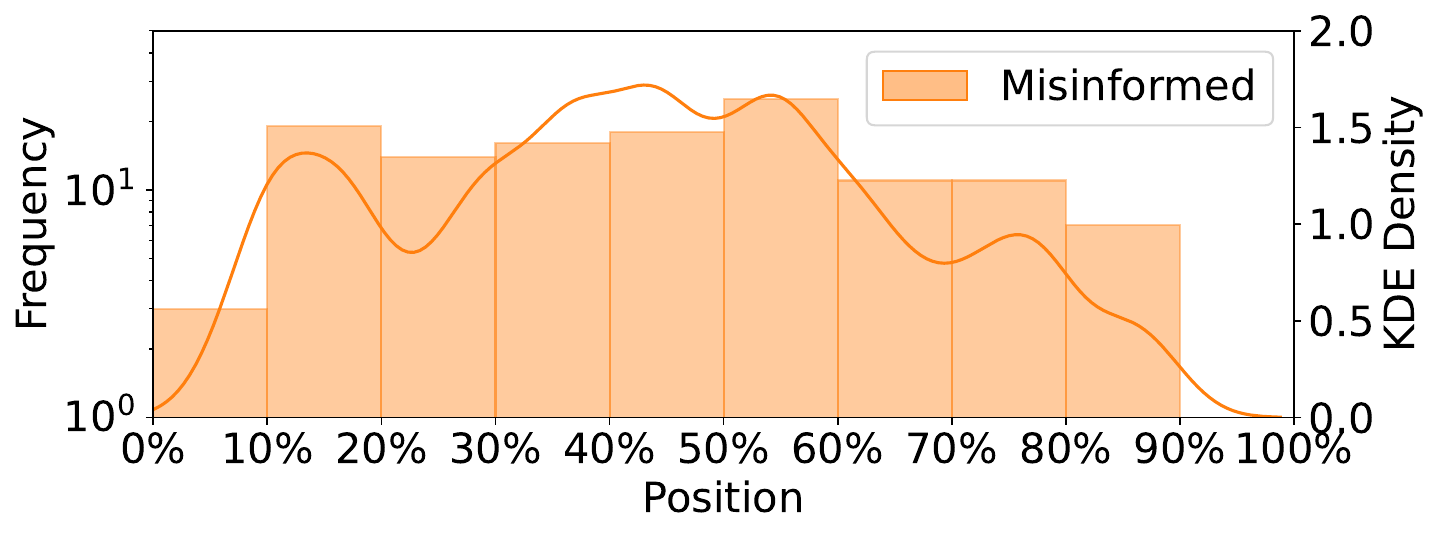}
  \end{center}
  \caption{Distribution of correction positions in the \textcolor{customorange}{misinformed} setting. Models identify misinformation at various positions.}
\label{fig:position_distribution}
\end{figure}

\subsection{Artificial Intelligence Assistant Usage} \label{sec:appendix:checklist:ai}
Artificial Intelligence (AI) assistants only aid us in AI code completions and grammar checking.

\SmallHeading{AI Code Completions}
During the development of the project, we used \href{https://github.com/features/copilot}{GitHub Copilot} to automatically complete inline and function header comments, some commonly used statements such as \texttt{with torch.no\_grad():}, as well as some repetitive functions or statements. We manually designed the structure of the code and implemented the basic logic of functions.

\SmallHeading{Grammar Checking}
During the paper writing, we manually wrote the draft paper and also used grammar checking tools, including \href{https://chatgpt.com/chat}{ChatGPT}, \href{https://www.grammarly.com/}{Grammarly}, and \href{https://www.deepl.com/translator}{DeepL} to avoid grammar mistakes.